\documentclass[11pt]{article}

\usepackage[final]{acl}

\usepackage{times}
\usepackage{latexsym}
\usepackage{microtype}
\usepackage{inconsolata}
\usepackage{graphicx}
\usepackage{booktabs}
\usepackage{tabularx}
\usepackage{amsmath,amssymb}
\usepackage{forest}
\usepackage{multirow}
\usepackage{pifont}
\useforestlibrary{edges}

\title{From Parameters to Behaviors: A Survey of Model Fusion for \\Large Language Models}

\author{
\textbf{Shuo Cai}\textsuperscript{1*},
\textbf{Yanggan Gu}\textsuperscript{1*},
\textbf{Zihao Wang}\textsuperscript{2*},
\textbf{Yuanyi Wang}\textsuperscript{1*},
\textbf{Yibo Yan}\textsuperscript{3},
\textbf{Wenjun Wang}\textsuperscript{1}
\\
\textbf{Yuhang Liu}\textsuperscript{5},
\textbf{Guanghao Zhu}\textsuperscript{1},
\textbf{Sirui Huang}\textsuperscript{1},
\textbf{Ming Li}\textsuperscript{1,4\textdagger},
\textbf{Hongxia Yang}\textsuperscript{1,4,5\textdagger}
\\
\textsuperscript{1}The Hong Kong Polytechnic University \\
\textsuperscript{2}The Chinese University of Hong Kong \\
\textsuperscript{3}The Hong Kong University of Science and Technology (Guangzhou) \\
\textsuperscript{4}Research Institute of Generative Artificial Intelligence, \textsuperscript{5}InfiX.ai \\
\texttt{shuo1031.cai@connect.polyu.hk}
}

\begin{document}
\sloppy
\maketitle
\begingroup
\renewcommand{\thefootnote}{}
\footnotetext{\textsuperscript{*}Equal contribution. \quad \textsuperscript{\textdagger}Corresponding authors.}
\endgroup

\begin{abstract}
Model fusion integrates the capabilities from source models into a single target model.
As of June 2026, Hugging Face hosts more than 2M models.
This growing pool provides a rich base for model reuse and capability integration.
Yet existing surveys often cover only separate parts of this space, and they do not provide a unified definition or a systematic taxonomy.
This survey defines model fusion and organizes prior work into three levels: parameter-level, representation-level, and behavior-level fusion.
We also review related metrics, benchmarks, and applications, summarize current challenges, and identify future directions.
Our goal is to provide a clear map of this area and support future work on model fusion.
A comprehensive list of papers about model fusion is available at \url{https://github.com/Baicaihaochi/Awesome-Model-Fusion-Survey}.
\end{abstract}

\section{Introduction}

As large language models and the open-source ecosystem continue to grow, the number and variety of available models have increased quickly.
As of June 2026, Hugging Face hosts more than 2M models\footnote{\url{https://huggingface.co/blog/huggingface/state-of-os-hf-spring-2026}}, providing a rich and diverse base for model reuse and capability integration.
Therefore, reusing and integrating existing model capabilities within a single model is becoming an important direction \citep{11268959,zheng2023learnmodel}.

\begin{figure}[t]
\centering
\includegraphics[width=\linewidth]{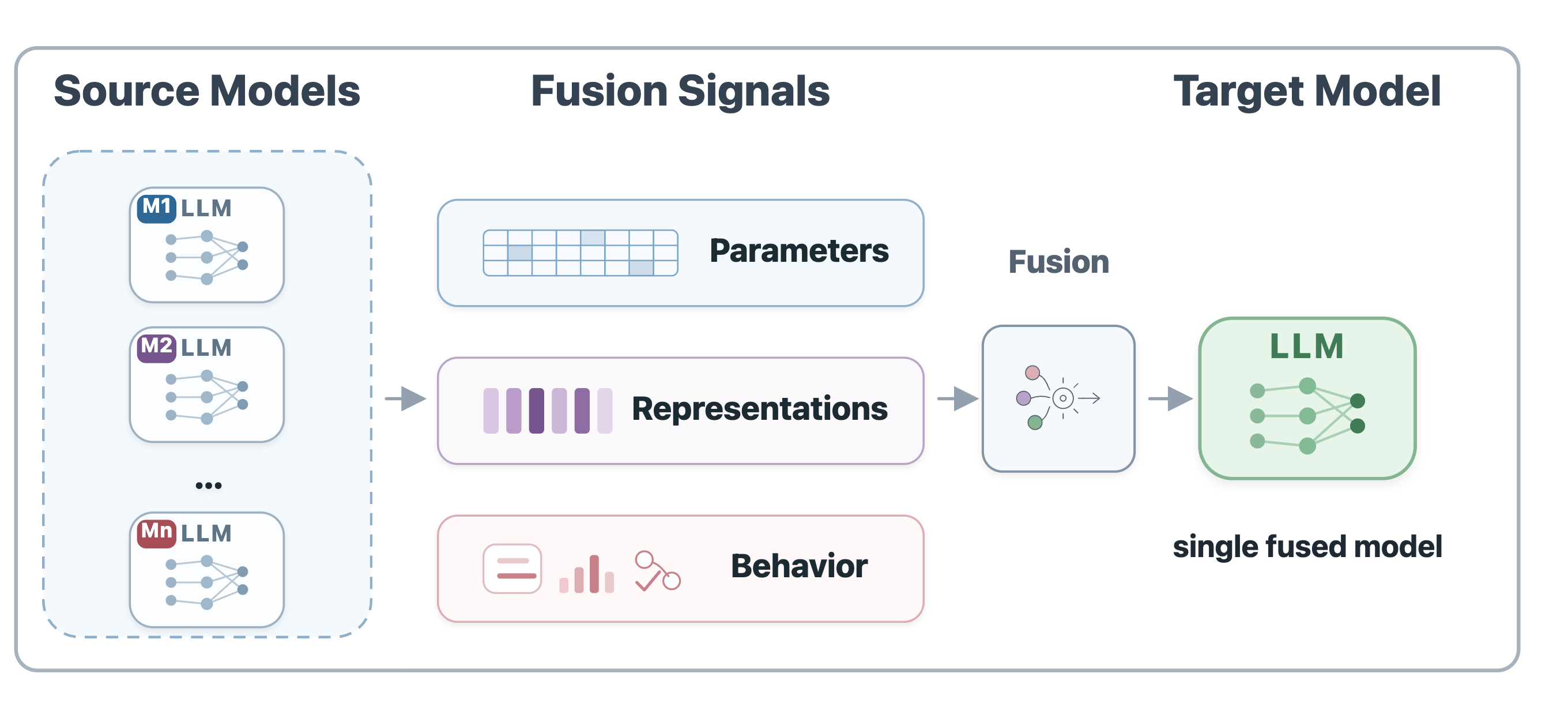}
\vspace{-2em}
\caption{Model fusion overview. Multiple source models contribute parameters, representations, or behaviors to construct one target model. After fusion, the target operates without the complete source models at inference time.}
\label{fig:model_fusion_demo}
\end{figure}

Given multiple source models with diverse capabilities, model fusion constructs a target model by combining their parameters, aligning their representations, or distilling their output behaviors, as shown in Figure~\ref{fig:model_fusion_demo}.
After fusion, the target model operates without relying on the complete source models at inference time.
Under this operational definition, traditional model merging and knowledge distillation can be viewed as parameter-level and behavior-level model fusion, respectively \citep{10.1145/3787849,song2026modelmergingeralarge,yadav2025surveymodel,11268959}.

\begin{figure*}[t]
\centering
\includegraphics[width=1.0\textwidth]{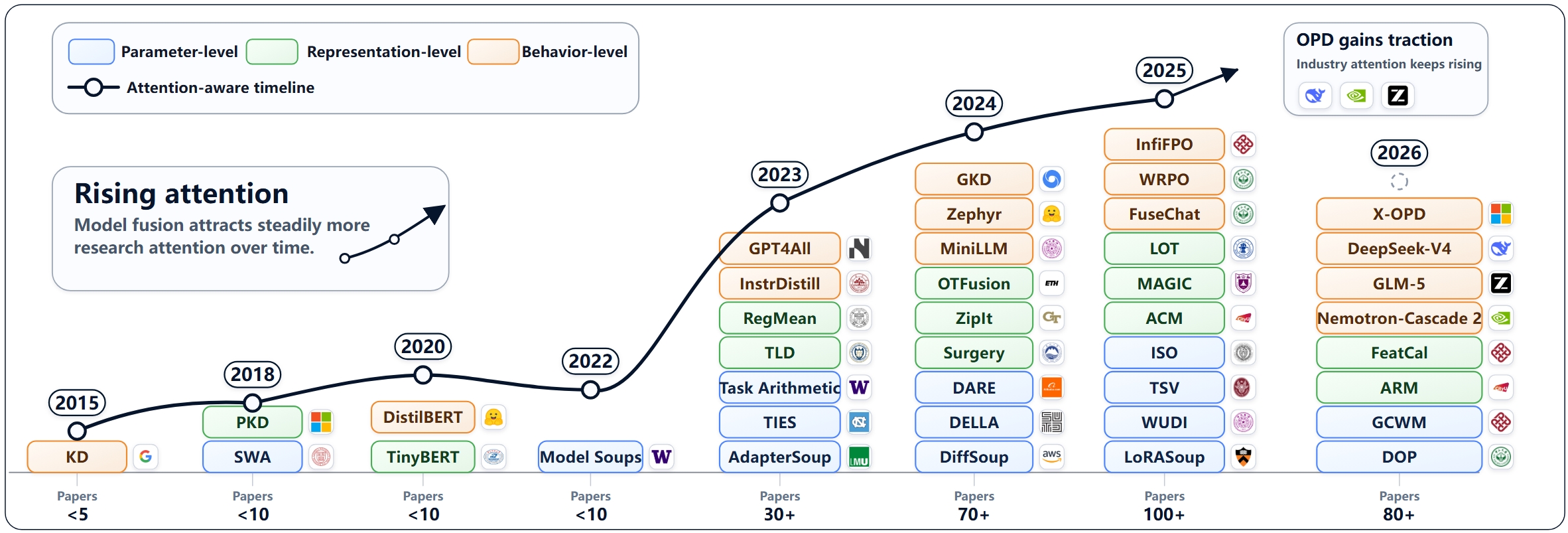}
\caption{Progress timeline of model fusion. Selected methods are color-coded by their primary fusion level. Annual counts indicate the broader growth trend; the 2026 count is a partial-year snapshot.}
\label{fig:model_fusion_progress}
\end{figure*}

Model fusion offers two practical advantages.
First, it enables efficient reuse of existing models and integrates their capabilities into one target model. This can be done by fusing parameters, using representations to diagnose and repair drift, or distilling output behaviors \citep{yadav2023tiesmerging,yang2024representationsurgery,wan2024knowledgefusion,agarwal2024policydistillation}. As shown in Figure~\ref{fig:model_fusion_progress}, model fusion has attracted steadily increasing researcher attention since 2023. This trend is also reflected in industrial practice, where DeepSeek-V4 \citep{deepseekai2026deepseekv4}, NVIDIA's Nemotron-Cascade 2 \citep{yang2026nemotroncascade}, and GLM-5 \citep{glm5team2026glm5} adopt on-policy distillation to integrate or recover model capabilities.
Second, model fusion supports continual learning by absorbing new task signals while preserving earlier capabilities. AIMMerging \citep{feng2025aimmergingadaptive}, RECALL \citep{wang2025recallrepresentation}, and NUFILT \citep{qiu2025nullspace} reduce forgetting.

Despite these advantages, recent studies also show that model fusion remains far from settled.
Weight averaging and alignment can improve accuracy and robustness, but some task-level combinations may collapse, and current theory cannot yet predict when fusion will succeed \citep{wortsman2022modelsoups,ainsworth2023gitre,cao2026empiricalstudy}.
Moreover, fusion becomes harder when source models differ in architecture, tokenizer, or modality, because parameter and representation alignment can be unstable \citep{sung2023empiricalstudy,cui2026transportmerge,du2025adammsmodel}.
In evaluation, recent benchmarks improve standardization, but average scores can still hide local degradation and cross-capability interference \citep{tang2025fusionbenchcomprehensive,he2025mergebenchbenchmark,tam2024realisticevaluation,cao2026empiricalstudy}.

\begin{table}[t]
\centering
\tiny
\setlength{\tabcolsep}{1.8pt}
\renewcommand{\arraystretch}{0.9}
\begin{tabularx}{\linewidth}{@{}>{\raggedright\arraybackslash}p{0.26\linewidth}>{\centering\arraybackslash}p{0.17\linewidth}>{\centering\arraybackslash}p{0.17\linewidth}>{\centering\arraybackslash}p{0.17\linewidth}>{\centering\arraybackslash}X@{}}
\toprule
Survey & Venue \& Year & Param. level & Repre. level & Behav. level \\
\midrule
\mbox{\citet{10.1007/s11263-021-01453-z}} & IJCV'21 &  & \(\checkmark\) & \(\checkmark\) \\
\mbox{\citet{xu2024surveyknowledge}} & arXiv'24 &  & \(\checkmark\) & \(\checkmark\) \\
\mbox{\citet{yadav2025surveymodel}} & TMLR'25 & \(\checkmark\) &  &  \\
\mbox{\citet{10.1145/3699518}} & TIST'25 &  & \(\checkmark\) & \(\checkmark\) \\
\mbox{\citet{Qin_2025}} & IJIS'25 & \(\checkmark\) &  & \(\checkmark\) \\
\mbox{\citet{10.1145/3787849}} & CSUR'26 & \(\checkmark\) & \(\checkmark\) &  \\
\mbox{\citet{song2026modelmergingeralarge}} & arXiv'26 & \(\checkmark\) & \(\checkmark\) &  \\
\mbox{\citet{11268959}} & TNNLS'26 & \(\checkmark\) & \(\checkmark\) &  \\
\mbox{\citet{song2026surveyonpolicydistillationlarge}} & arXiv'26 &  &  & \(\checkmark\) \\
\mbox{\citet{fang2026knowledgedistillationdatasetdistillation}} & AIR'26 &  & \(\checkmark\) & \(\checkmark\) \\
\midrule
\textbf{Ours} &  & \(\checkmark\) & \(\checkmark\) & \(\checkmark\) \\
\bottomrule
\end{tabularx}
\caption{Coverage of related surveys across the three fusion levels. A checkmark indicates substantive treatment of methods at that level; only our survey covers parameter-, representation-, and behavior-level fusion together.}
\label{tab:related_survey_comparison}
\end{table}

As shown in Table~\ref{tab:related_survey_comparison}, existing surveys treat model merging and knowledge transfer separately \citep{xu2024surveyknowledge,song2026surveyonpolicydistillationlarge,10.1007/s11263-021-01453-z,10.1145/3699518,Qin_2025,fang2026knowledgedistillationdatasetdistillation}, without a unified view of the scope, boundary, and evaluation of model fusion.
We address this gap with a formal definition and a three-level taxonomy, followed by evaluation, applications, and open research problems.

We survey more than 150 papers. Section~2 defines model fusion, and Section~3 presents the three-level taxonomy and evaluation settings. Sections~4--6 present practical takeaways and challenges.

\section{Definition and Formulation}

\paragraph{Definition.}
We define model fusion as follows:
\begin{quote}
Model fusion constructs a target model by combining parameters, aligning representations, or distilling output behaviors from multiple source models. Its goal is to integrate the knowledge and capabilities carried by the source models while retaining efficient inference.
\end{quote}

Accordingly, we include a method as model fusion only if its construction of the target model uses at least one of these three operations and the resulting target model can operate at inference time without relying on the complete source models.
Whether an operation occurs during training or after training is orthogonal to this definition: model merging is often training-free, whereas representation alignment and behavior distillation usually optimize the target model during fusion.
Task Arithmetic combines parameter updates, Representation Surgery aligns hidden representations, and FuseLLM distills output distributions; each produces a target that is independent of the complete source models at inference time \citep{ilharco2023editing,yang2024representationsurgery,wan2024knowledgefusion}.

These criteria exclude inference-time model combination and selection.
Ensembles aggregate predictions from multiple models, while routing systems such as RouteLLM select a source model for each query; both retain and invoke source models at inference time rather than constructing an inference-independent target \citep{chen2026harnessingmultiplelargelanguage,ong2025routellm}.
LLM-as-a-Judge is also excluded because it evaluates model outputs rather than constructing a target model through parameter combination, representation alignment, or behavior distillation \citep{zheng2023judgingllm}.
We further distinguish behavior distillation from reward-only reinforcement learning.
On-policy distillation qualifies when a teacher provides dense, token-level behavioral supervision, such as output-distribution targets or an equivalent KL-constrained signal, on target-generated states \citep{agarwal2024policydistillation,yang2026learningbeyond}; using an LLM only for a scalar reward does not qualify because it does not distill the teacher's output behavior.

Let \(\mathcal X\) and \(\mathcal Y\) be the input space and the output space.
Given \(n\) source models
\begin{equation}
\label{eq:source-model-set}
\mathcal S=\{M_i^{\mathrm{src}}\}_{i=1}^{n},
\end{equation}
where \(M_i^{\mathrm{src}}\) is the \(i\)-th source model.
For input \(x\in\mathcal X\), each source model gives a conditional output distribution \(p_i^{\mathrm{src}}(y\mid x)\), where \(y\in\mathcal Y\).
The goal is to build a target model \(M_{\theta}^{\mathrm{tgt}}\) with parameters \(\theta\).
Its conditional output distribution is \(p_{\theta}^{\mathrm{tgt}}(y\mid x)\).
Model fusion can be written as a mapping from source models to the target model:
\begin{equation}
\label{eq:fusion-mapping}
\theta=\Phi(\mathcal S,\mathcal D) .
\end{equation}
Here, \(\Phi\) is the fusion mapping.
\(\mathcal D\) denotes optional fusion data and can be empty.
The general fusion goal can be expressed as
\begin{equation}
\label{eq:general-fusion-objective}
\begin{split}
\theta^\star
&=\arg\min_\theta
\sum_{i=1}^{n}\mathbb E_{x\sim\mathcal T_i}\Bigl[\\[-0.2em]
&\quad
D_{\mathrm{out}}\!\left(
p_i^{\mathrm{src}}(\cdot\mid x),
p_\theta^{\mathrm{tgt}}(\cdot\mid x)
\right)\Bigr].
\end{split}
\end{equation}
Here, \(\theta^\star\) denotes the fused target parameters, \(\mathcal T_i\) is the task distribution of source \(i\), and \(D_{\mathrm{out}}\) measures the discrepancy between the source and target output distributions.
Equation~\ref{eq:general-fusion-objective} states a general retention goal rather than a shared training loss; a fusion method need not optimize it explicitly.

\paragraph{Inference Independence.}
Once \(\theta\) is fixed, the target model no longer needs the source models during inference:
\[
p_{\theta}^{\mathrm{tgt}}(y\mid x,\mathcal S)
=
p_{\theta}^{\mathrm{tgt}}(y\mid x).
\]

\section{Taxonomy of Model Fusion}

\definecolor{mc1}{RGB}{109,173,209}
\definecolor{mc2}{RGB}{182,215,232}
\definecolor{mc3}{RGB}{233,241,244}
\definecolor{mcleaf}{RGB}{251,227,213}
\definecolor{param}{HTML}{DAE8FC}
\definecolor{repr}{HTML}{E1D5E7}
\definecolor{beh}{HTML}{F8CECC}
\definecolor{eval}{HTML}{E7EEF8}
\definecolor{fusionroot}{HTML}{F9F7ED}

\tikzstyle{my-box}=[
    rectangle,
    rounded corners,
    text opacity=1,
    minimum height=1.5em,
    minimum width=6em,
    inner sep=3pt,
    align=center,
    fill opacity=.3,
    thick,
]
\tikzstyle{leaf}=[
    my-box,
    fill=gray!3,
    text=black,
    align=left,
    text width=30.5em,
    font=\normalsize,
    inner xsep=3pt,
    inner ysep=4pt,
]

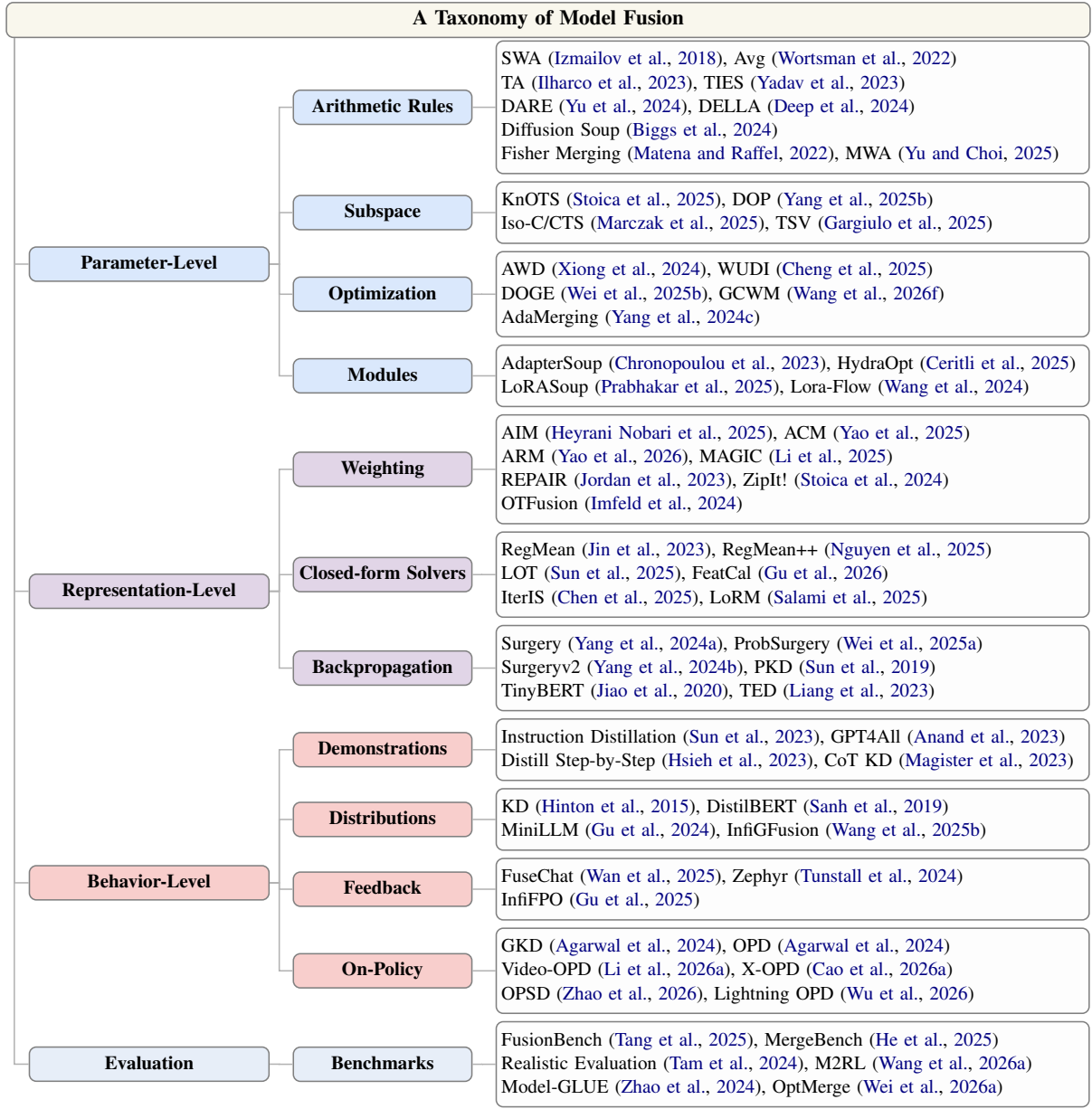
\begin{figure*}[!t]
    \centering
    \resizebox{\textwidth}{!}{
        \begin{forest}
            forked edges,
            for tree={
                grow=east,
                reversed=true,
                anchor=base west,
                parent anchor=east,
                child anchor=west,
                base=center,
                font=\normalsize,
                rectangle,
                draw=gray,
                rounded corners,
                align=left,
                text centered,
                minimum width=5em,
                edge+={gray!60, thick},
                s sep=4pt,
                inner xsep=3pt,
                inner ysep=3pt,
                thick,
                fit=band,
            },
            where level=0{
                folder,
                grow'=0,
                fill=fusionroot,
                font=\large,
                text width=56.2em,
                child anchor=west,
                parent anchor=south west,
                anchor=west,
                align=center,
                calign=first,
                yshift=105pt,
            }{},
            where level=1{text width=12em}{},
            where level=2{text width=8.8em}{},
            where level=3{text width=15em}{},
            [
                {
                    \textbf{A Taxonomy of Model Fusion}
                    % \quad
                    % \resizebox{0.64\linewidth}{!}{%
                    % $\mathcal S=\{M_i^{\mathrm{src}}\}_{i=1}^{n}
                    % \quad
                    % \theta^\star=\Phi(\mathcal S,\mathcal D)=\theta^\star\!=\!\arg\min_{\theta}\sum_{i=1}^{n}\mathop{\mathbb E}\limits_{x\sim\mathcal T_i}\![\mathrm{D}_{\mathrm{out}}(p_{\theta}^{\mathrm{tgt}}(\cdot\mid x),p_i^{\mathrm{src}}(\cdot\mid x))].
                    % \quad
                    % p_{\theta}^{\mathrm{tgt}}(y\mid x,\mathcal S)=p_{\theta}^{\mathrm{tgt}}(y\mid x)$}
                }
                [
                    {\textbf{Parameter-Level}
                    % \\
                    % {\scriptsize$\theta=\Phi^{\mathrm{param}}(\mathcal S)$}
                    }
                    , fill=param, align=center
                    [
                        \textbf{Arithmetic Rules}, fill=param
                        [
                            {SWA~\citep{izmailov2018averagingweights}, Avg~\citep{wortsman2022modelsoups}\\
                            TA~\citep{ilharco2023editing}, TIES~\citep{yadav2023tiesmerging}\\
                            DARE~\citep{yu2024languagemodels}, DELLA~\citep{deep2024dellamerging}\\
                            Diffusion Soup~\citep{biggs2024diffusionsoup}\\
                            Fisher Merging~\citep{matena2022mergingmodels}, MWA~\citep{yu2025parameterefficient}}
                            , leaf
                        ]
                    ]
                    [
                        \textbf{Subspace}, fill=param
                        [
                            {KnOTS~\citep{stoica2024modelmerging}, DOP~\citep{yang2026continual}\\
                            Iso-C/CTS~\citep{marczak2025notask}, TSV~\citep{gargiulo2025tasksingular}\\
                            }
                            , leaf
                        ]
                    ]
                    [
                        \textbf{Optimization}, fill=param
                        [
                            {AWD~\citep{xiong2024multitask}, WUDI~\citep{cheng2025whoever}\\
                            DOGE~\citep{wei2025modelingmulti}, GCWM~\citep{wang2026geometry}\\
                            AdaMerging~\citep{yang2024adamerging}}
                            , leaf
                        ]
                    ]
                    [
                        \textbf{Modules}, fill=param
                        [
                            {AdapterSoup~\citep{chronopoulou2023adaptersoupweight}, HydraOpt~\citep{ceritli-etal-2025-hydraopt}\\
                            LoRASoup~\citep{prabhakar2025lorasoups}, Lora-Flow~\citep{wang2024loraflow}}
                            , leaf
                        ]
                    ]
                ]
                [
                    {\textbf{Representation-Level}
                    % \\
                    % \resizebox{0.95\linewidth}{!}{$\theta^\star=\arg\min_{\theta}\sum_{i=1}^{n}\sum_{\ell}\mathop{\mathbb E}\limits_{x\sim\mathcal T_i}\bigl[\mathrm D_{\mathrm{out}}\bigl(r_{\theta}^{\ell}(\cdot\mid x),r_i^{\ell}(\cdot\mid x)\bigr)\bigr]$}
                    }
                    , fill=repr, align=center
                    [
                        \textbf{Weighting}, fill=repr
                        [
                            {AIM~\citep{nobari2025activationinformed}, ACM~\citep{yao2025activationguided}\\
                            ARM~\citep{yao2026mergingbeyond}, MAGIC~\citep{li2025magicachieving}\\
                            REPAIR~\citep{jordan2023repair}, ZipIt!~\citep{stoica2024zipitmerging}\\
                            OTFusion~\citep{imfeld2024transformerfusion}}
                            , leaf
                        ]
                    ]
                    [
                        \textbf{Closed-form Solvers}, fill=repr
                        [
                            {RegMean~\citep{jin2023datalessknowledge}, RegMean++~\citep{nguyen2025regmeanenhancing}\\
                            LOT~\citep{sun2025towardsminimizing}, FeatCal~\citep{gu2026featcalfeaturecalibrationpostmerging}\\
                            IterIS~\citep{chen2025iterisiterative}, LoRM~\citep{salami2025closedform}}
                            , leaf
                        ]
                    ]
                    [
                        \textbf{Backpropagation}, fill=repr
                        [
                            {Surgery~\citep{yang2024representationsurgery}, ProbSurgery~\citep{wei2025representationsurgery}\\
                            Surgeryv2~\citep{yang2024surgeryv2bridging}, PKD~\citep{sun2019patientknowledge}\\
                            TinyBERT~\citep{jiao2020tinybertdistilling}, TED~\citep{liang2023lessmore}}
                            , leaf
                        ]
                    ]
                ]
                [
                    {\textbf{Behavior-Level}
                    \\
                    % \resizebox{0.95\linewidth}{!}{$\theta^\star=\arg\min_{\theta}\sum_{i=1}^{n}\mathop{\mathbb E}\limits_{x\sim\mathcal T_i}\Bigl[\mathrm D_{\mathrm{out}}\bigl(q_{\theta}^{x},q_i^{x}\bigr)\Bigr]$}
                    }
                    , fill=beh, align=center
                    [
                        \textbf{Demonstrations}, fill=beh
                        [
                            {Instruction Distillation~\citep{sun2023instructiondistillation}, GPT4All~\citep{anand2023gpt4alltraining}\\
                            Distill Step-by-Step~\citep{hsieh2023distillingstep}, CoT KD~\citep{magister2023teachingsmall}}
                            , leaf
                        ]
                    ]
                    [
                        \textbf{Distributions}, fill=beh
                        [
                            {KD~\citep{hinton2015distillingknowledge}, DistilBERT~\citep{sanh2019distilbertdistilled}\\
                            MiniLLM~\citep{gu2024minillmknowledge},
                            InfiGFusion~\citep{wang2026infigfusion}}
                            , leaf
                        ]
                    ]
                    [
                        \textbf{Feedback}, fill=beh
                        [
                            {FuseChat~\citep{wan2024fusechatknowledge}, Zephyr~\citep{tunstall2023zephyrdirect}\\
                            InfiFPO~\citep{gu2026infifpo}}
                            , leaf
                        ]
                    ]
                    [
                        \textbf{On-Policy}, fill=beh
                        [
                            {GKD~\citep{agarwal2024policydistillation}, OPD~\citep{agarwal2024policydistillation}\\
                            Video-OPD~\citep{li2026videoopd}, X-OPD~\citep{cao2026xopd}\\
                            OPSD~\citep{zhao2026selfdistilled},
                            Lightning OPD~\citep{wu2026lightning}}
                            , leaf
                        ]
                    ]
                ]
                [
                    {\textbf{Evaluation}}
                    , fill=eval, align=center, calign=child, calign child=1, yshift=13pt
                    [
                        \textbf{Benchmarks}, fill=eval, align=center, tier=evalbench
                        [
                            {FusionBench~\citep{tang2025fusionbenchcomprehensive}, MergeBench~\citep{he2025mergebenchbenchmark}\\
                            Realistic Evaluation~\citep{tam2024realisticevaluation}, M2RL~\citep{wang2026mixmerge}\\
                            Model-GLUE~\citep{zhao2024modelglue}, OptMerge~\citep{wei2025optmergeunifying}}
                            , leaf
                        ]
                    ]
                ]
            ]
        \end{forest}
    }
    \caption{A taxonomy of model fusion for LLMs and MLLMs. The method branches are organized by the main object being fused or aligned: parameters, representations, or behaviors. AdaMerging, MWA, and Fisher Merging remain parameter-level because their learned or estimated quantities determine parameter combinations rather than provide external source behavior. The evaluation branch summarizes representative benchmark resources. Methods and resources are illustrative rather than exhaustive.}
    \label{fig:model-fusion-overview}
\end{figure*}

We organize methods by the source signal used in fusion rather than the surface form of the final result: parameters, representations, or behaviors.

\subsection{Parameter-Level Fusion}

\textbf{Definition.}
Let \(\theta_0\) be the reference parameters and, for each source \(M_i^{\mathrm{src}}\in\mathcal S\), let \(\theta_i\) be its parameters and \(\Delta_i=\theta_i-\theta_0\) its update.
Parameter-level fusion transforms and combines these updates:
\begin{equation}
\label{eq:parameter-fusion-mapping}
\Phi_{\mathrm P}(\mathcal S,\mathcal D)
=\theta_0+\sum_{i=1}^{n}\alpha_i A_i(\Delta_i).
\end{equation}
Here, \(\alpha_i\in\mathbb R\) is a scalar source coefficient, and \(A_i\) can be the identity, masking or rescaling, a permutation, or a subspace projection; both can be determined without data or estimated from \(\mathcal D\).

\textbf{Related work and methods.}
Parameter-level fusion methods can be organized by how they manipulate parameters.
%\emph{Arithmetic rules} are the most basic merging methods, typically avoiding explicit training or complex objectives and combining model capabilities through direct computations in parameter space.
\emph{Arithmetic rules} combine source weights or parameter deltas with fixed or lightly tuned coefficients.
SWA \citep{izmailov2018averagingweights} averages parameter snapshots sampled along an SGD trajectory with a cyclical or constant learning rate, approximating an ensemble with a single model and improving generalization with little additional cost.
Model soups \citep{wortsman2022modelsoups} show that multiple fine-tuned models can be averaged when they lie in a nearby parameter basin, while task arithmetic \citep{ilharco2023editing} represents the difference between a fine-tuned model and its base model as a task vector, enabling capability composition or behavior editing through vector addition and subtraction.
Fisher Merging weights individual parameters using sample-estimated Fisher information, while MWA weights checkpoints using training metrics such as loss or training step \citep{matena2022mergingmodels,yu2025parameterefficient}.
Subsequent methods further address conflicts and redundancy among parameter deltas.
For example, TIES-Merging, DARE, and DELLA-Merging reduce interference through sign consistency, random dropping or rescaling \citep{yadav2023tiesmerging,yu2024languagemodels,deep2024dellamerging}.

\emph{Subspace methods} identify, reshape, or constrain structured directions in weight or update space to improve alignment and reduce interference.
SVD-based methods use singular directions to reshape update spaces, separate shared and task-specific components, and reduce interference \citep{stoica2024modelmerging,marczak2025notask,gargiulo2025tasksingular}.
% SVD-based methods decompose task matrices into singular directions, reshape singular spectra, separate common and task-specific subspaces, or decorrelate task singular vectors \citep{stoica2024modelmerging,marczak2025notask,gargiulo2025tasksingular}.
For continual fusion, DOP \citep{yang2026continual} approximates unavailable data subspaces with SVD subspaces of task vectors and applies dual orthogonal projections to balance stability and plasticity without accessing task data.

\emph{Optimization methods} formulate merge coefficients, task vectors, or transformation variables as explicit optimization problems.
AdaMerging learns task- or layer-wise fusion coefficients by minimizing output entropy on unlabeled data; the resulting coefficients still combine source parameters rather than distilling source behaviors \citep{yang2024adamerging}.
AWD \citep{xiong2024multitask} optimizes a decomposition of task vectors into redundant and disentangled components, improving orthogonality while preserving task-specific performance.
WUDI \citep{cheng2025whoever} uses task vectors to locate interference sources and correct the responsible components without data.
GCWM \citep{wang2026geometry} and DOGE \citep{wei2025modelingmulti} use geometric or projected-gradient objectives to reduce interference during multi-task fusion.
% GCWM \citep{wang2026geometry} uses state-relative geometry conflict and Wasserstein barycenters to gate update integration in continual post-training, while DOGE \citep{wei2025modelingmulti} formulates multi-task merging as adaptive projective gradient descent over task-vector corrections constrained by a shared subspace.

\emph{Module fusion} combines LoRA, adapters, projectors, or other pluggable modules instead of full-model parameters, suiting parameter-efficient fine-tuning.
AdapterSoup averages domain adapters, whereas LoRA soups average, concatenate, or weight skill-specific modules \citep{chronopoulou2023adaptersoupweight,hu2022loralow,prabhakar2025lorasoups}.
% to closed-form LoRA merging methods such as LoRM, which adapt RegMean-style objectives to solve for merged LoRA matrices that match the responses of client- or task-specific modules \citep{chronopoulou2023adaptersoupweight,hu2022loralow,prabhakar2025lorasoups,salami2025closedform}.

Parameter-level fusion is efficient but requires source compatibility and interference control.
% Overall, parameter-level fusion provides a direct route to deployable models with much lower inference cost than ensemble-based alternatives.
% Arithmetic rules are the simplest, but they rely heavily on source-model compatibility.
% Subspace and optimization methods better handle task interference by reshaping or searching over merge directions, often relying on parameter statistics, structural assumptions, or lightweight search.
% Although most methods in this group operate primarily on parameters, a few seemingly simple variants, such as model soups and AdapterSoup, may use held-out data to select candidate models, tune merge coefficients, or calibrate module combinations before deployment.
% Module merging is especially useful for lightweight adaptation in LLMs and MLLMs, though dynamic variants may retain extra gating components at inference time.
% The main limitations remain parameter alignment, negative transfer, and capability forgetting.
% Recent trends point toward fine-grained merge coefficients, module-level fusion, and tighter integration with representation diagnostics.

\begin{figure*}[t]
\centering
\includegraphics[width=\textwidth]{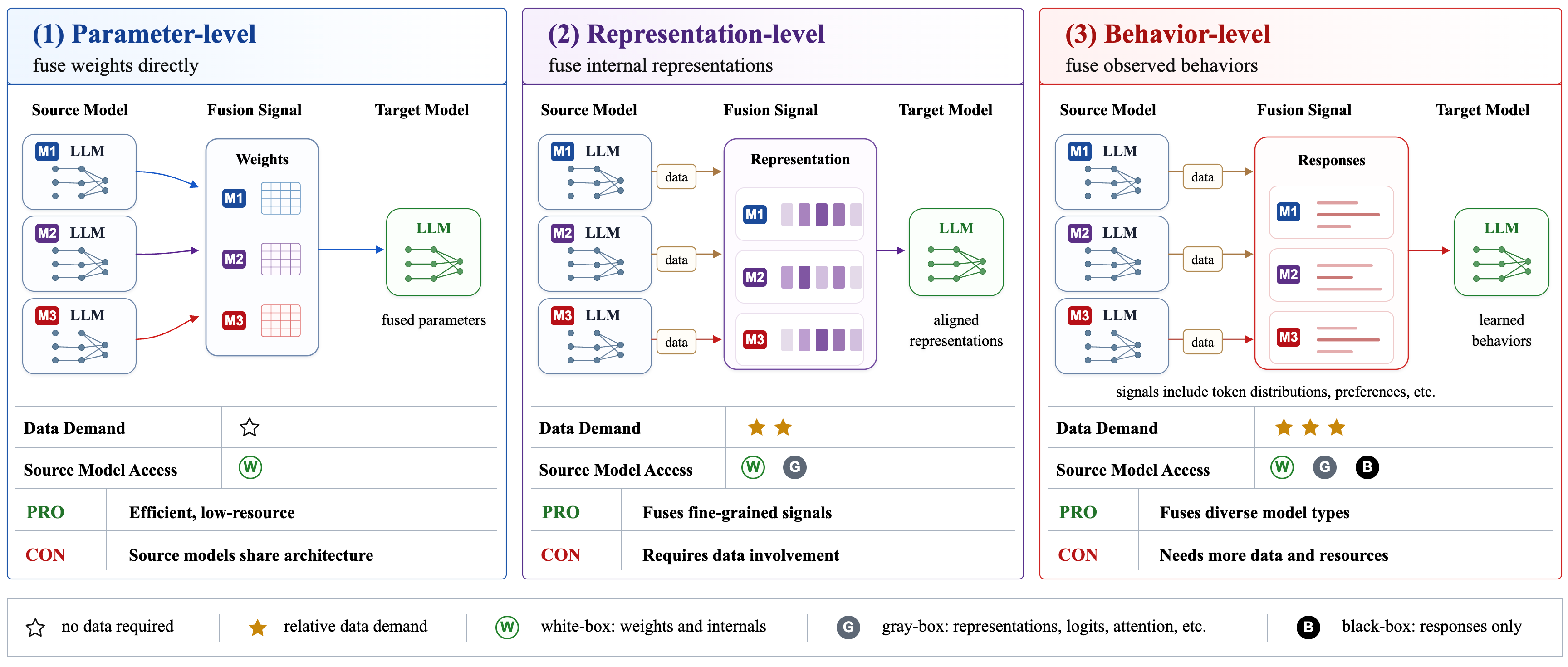}
\caption{Three levels of model fusion and their practical trade-offs. Data demand and source access indicate typical relative requirements rather than strict rules. A pipeline may combine levels, but its primary level is determined by the signal used to construct or update the target model.}
\label{fig:model-fusion-three-levels-latex}
\end{figure*}

\subsection{Representation-Level Fusion}

\textbf{Definition.}
Representation-level fusion uses intermediate representations as the main signal for capability integration.
For each \(M_i^{\mathrm{src}}\in\mathcal S\), let \(\widetilde r_i^{\ell}(x)\) denote its representation after any required layer matching, normalization, or dimensional projection, and let \(r_\theta^{\ell}(x)\) be the target representation, where \(\ell\in\mathcal L\) indexes the matched layers.
We write \(\mathcal D=(\mathcal D_i)_{i=1}^{n}\), where \(\mathcal D_i\) is the fusion-data distribution for source \(i\).
Representation-level fusion matches these signals as
\begin{equation}
\label{eq:representation-fusion-objective}
\begin{split}
\Phi_{\mathrm R}(\mathcal S,\mathcal D)
&=\arg\min_\theta
\sum_{\ell\in\mathcal L}\sum_{i=1}^{n}\\[-0.2em]
&\quad\mathbb E_{x\sim\mathcal D_i}\Bigl[
D_{\mathrm R}\!\left(\widetilde r_i^{\ell}(x),r_\theta^{\ell}(x)\right)
\Bigr].
\end{split}
\end{equation}
The discrepancy \(D_{\mathrm R}\) is instantiated according to the matched signal, such as \(L_1\) or MSE for feature tensors, cosine or correlation discrepancy for feature directions or unit matching, and MMD or moment discrepancy for representation distributions or activation calibration.
Representation statistics on \(\mathcal D_i\) can also determine \(\alpha_i\) or \(A_i\) in Equation~\ref{eq:parameter-fusion-mapping}; such methods remain representation-level when these statistics are the main fusion signal.

\textbf{Related work and methods.}
Representation-level fusion asks how intermediate representations can guide the construction or repair of a target model.
Existing methods mainly use representations in three ways: to derive merge signals, solve local matching problems, or train repair and distillation objectives.

\emph{Weighting methods} compute fusion weights from representations and then combine models in parameter space.
These weights can be defined over parameters, layers, modules, or matched components.
AIM \citep{nobari2025activationinformed} estimates weight saliency from activation magnitudes on a task-agnostic calibration set. MAGIC \citep{li2025magicachieving} calibrates representation and weight magnitudes, while Merging Beyond \citep{yao2026mergingbeyond} uses activation subspaces to form rotation-aware updates.
Related alignment methods compute correspondence from representations before fusion: REPAIR \citep{jordan2023repair} rescales preactivations, ZipIt! \citep{stoica2024zipitmerging} matches units by activation similarity, and Transformer Fusion \citep{imfeld2024transformerfusion} aligns Transformer components with optimal transport.
These methods are efficient, but they depend on calibration data, layer correspondence, and reliable representation similarity.

\emph{Closed-form solvers} formulate representation matching as local regression problems and solve them analytically, which is most practical for linear modules.
RegMean \citep{jin2023datalessknowledge} uses input covariance to merge each linear module so that its output matches source-module outputs. RegMean++ \citep{nguyen2025regmeanenhancing} improves this local view by adding intra-layer and cross-layer dependencies. LOT-Merging \citep{sun2025towardsminimizing} and FeatCal \citep{gu2026featcalfeaturecalibrationpostmerging} further treat representation drift as the main target: the former derives layer-wise analytic updates, while the latter calibrates merged weights in forward order by separating upstream propagation from local mismatch. For LoRA fusion, LoRM \citep{salami2025closedform} applies output matching to low-rank modules, and IterIS \citep{chen2025iterisiterative} refines the matching objective through iterative inference-solving.
Compared with weighting methods, these solvers use representations more directly by fitting local matching objectives, not only by estimating fusion weights.

\emph{Backpropagation methods} train the target model or added repair modules with representation losses, allowing nonlinear repair and the use of multiple internal signals such as hidden states and attention maps.
Patient Knowledge Distillation \citep{sun2019patientknowledge} augments output distillation with hidden-state matching at selected source layers, while TinyBERT \citep{jiao2020tinybertdistilling} combines prediction-layer supervision with matching of embeddings, attention maps, and hidden states.
TED \citep{liang2023lessmore} likewise augments output distillation with task-aware filters that retain task-relevant hidden representations before alignment.
Fusion repair methods use the same idea after an initial parameter-level fusion step. Representation Surgery \citep{yang2024representationsurgery} learns a lightweight module to correct final-layer representation bias. Surgeryv2 \citep{yang2024surgeryv2bridging} extends this repair across multiple layers. ProbSurgery \citep{wei2025representationsurgery} models the correction as a distribution to capture uncertainty from parameter interference. Compared with closed-form solvers, these methods can handle more complex mismatch, but they require slower training, larger repair data, and careful regularization to avoid overfitting.

Representation-level fusion is most useful when hidden states expose drift or layer mismatch. Weighting methods are inexpensive but sensitive to calibration; closed-form solvers need aligned linear modules and more representation samples; backpropagation handles complex mismatch but requires more training and data. Open problems include efficient drift repair and alignment across dissimilar models.

\subsection{Behavior-Level Fusion}

\textbf{Definition.}
Behavior-level fusion uses observable source behaviors to train a target model.
Let \(s\) denote an input state, such as \((x,y_{<t})\) for an autoregressive model, and let \(\mu_{\mathcal D}\) be the empirical state distribution from fixed or target-generated sequences.
Let \(q_{\mathcal S}(\cdot\mid s)\) be a behavior target constructed by selecting, aggregating, or aligning source outputs.
Output-distribution distillation uses
\begin{equation}
\label{eq:behavior-fusion-objective}
\begin{split}
\Phi_{\mathrm B}(\mathcal S,\mathcal D)
&=\arg\min_\theta
\mathbb E_{s\sim\mu_{\mathcal D}}\Bigl[\\[-0.2em]
&\quad
D_{\mathrm B}\!\left(
q_{\mathcal S}(\cdot\mid s),
p_\theta^{\mathrm{tgt}}(\cdot\mid s)
\right)\Bigr].
\end{split}
\end{equation}
Here, \(p_\theta^{\mathrm{tgt}}(\cdot\mid s)\) is the target output distribution.
Common choices for \(D_{\mathrm B}\) include forward KL, reverse KL, and generalized JSD.
Source models therefore act as \emph{behavior providers}, rather than parameter or representation providers. Methods using only target-model entropy, confidence, or uncertainty for parameter fusion lack external behavioral supervision and fall outside this category.

\textbf{Related work and methods.}
Behavior-level fusion can be grouped by the transferred behavior type into distribution fusion, demonstration fusion, and feedback fusion, with an orthogonal distinction between off-policy supervision on fixed data and on-policy supervision on target-induced states.

\emph{Distribution fusion} transfers source-provided soft labels, output distributions, token probabilities, or logits. Classical knowledge distillation matches a teacher's softened output distribution \citep{hinton2015distillingknowledge}, while DistilBERT shows its effectiveness for language model compression \citep{sanh2019distilbertdistilled}. For model fusion, such signals can integrate complementary capabilities across models: InfiGFusion further models logits as relational graphs and aligns their geometry via an efficient Gromov--Wasserstein approximation, moving beyond independent token-level matching \citep{wang2026infigfusion}. 
\emph{Demonstration fusion} learns from source-generated responses, rationales, reasoning traces, tool-use traces, or trajectories. Instruction distillation and GPT4All-style training use stronger-model outputs to train independently deployable targets \citep{sun2023instructiondistillation,anand2023gpt4alltraining}, while rationale or step-level distillation transfers intermediate reasoning processes \citep{hsieh2023distillingstep,magister2023teachingsmall}. These methods require only sampled outputs, but may inherit source errors, spurious reasoning, or stylistic bias.

\emph{Feedback fusion} transfers preferences, scores, critiques, corrections, reward signals, or verifier labels, making it useful for alignment and safety transfer when parameters, hidden states, or full distributions are unavailable. Chat-oriented fusion can construct data from multi-source responses, rankings, and preferences, as in FuseChat and Zephyr \citep{wan2024fusechatknowledge,tunstall2023zephyrdirect}. InfiFPO further formulates fusion as implicit preference optimization, absorbing source-model advantages without direct pivot model access \citep{gu2026infifpo}. Source-derived preferences, critiques, or verifier feedback qualify when they transfer identifiable source behavior, whereas generic reward-only RL without such behavioral transfer is excluded.

From the state-distribution perspective, \emph{off-policy fusion} uses fixed behavior data and is simple to scale, but suffers from mismatch when the target visits poorly covered states. \emph{On-policy fusion} instead lets the target generate prefixes, responses, or trajectories, and then obtains supervision on these target-induced states. This connects to dataset aggregation in imitation learning \citep{ross2011reductionimitation}; in LLMs, GKD instantiates it by distilling from teacher feedback on student-generated sequences \citep{agarwal2024policydistillation}. Recent OPD variants study self-distillation, black-box or semi-on-policy supervision, offline logit reuse, token-efficient supervision, and stabilization \citep{zhao2026selfdistilled,chen2026soda,wu2026lightning,xu2026tip,luo2026demystifying}, and extend OPD to multimodal trajectories such as video grounding and speech LLM alignment \citep{li2026videoopd,cao2026xopd}. This formulation is especially suitable for heterogeneous fusion, where source and target models may differ in architecture, tokenizer, modality interface, decoding policy, or capability profile.

Behavior-level fusion suits closed-source and heterogeneous models because it avoids parameter and hidden-state access. Demonstrations may carry imitation bias; distributions often require logit access; feedback depends on verifier or reward quality. Open problems include robust multi-source aggregation, budget-aware on-policy queries, and joint process- and outcome-level feedback.

\subsection{Evaluation}
\label{sec:eval}

\paragraph{Metrics.}
Evaluation for model fusion can start from two simple metrics.
(1) \textit{Avg performance} reports the average performance of the target model over the task pool.
It gives a direct view of overall quality and is easy to compare across methods.
(2) \textit{Normalized performance} compares the target model with the corresponding source model on each task.
MergeBench uses this metric to measure how much source task performance is retained by the target model \citep{he2025mergebenchbenchmark}.
This is important because a target model can improve the average score while losing one source capability.
Other metrics cover interference, generalization, alignment, cost, and safety when applicable.
Appendix Table~\ref{tab:fusion_metrics} summarizes these metrics.
% Table~\ref{tab:avg_score_limit} illustrates this point with MergeBench results.
% Methods with close average scores can retain source capabilities to different degrees and show very different worst-task drops.

% \begin{table}[t]
% \centering
% \tiny
% \setlength{\tabcolsep}{2.5pt}
% \renewcommand{\arraystretch}{0.96}
% \resizebox{\linewidth}{!}{
% \begin{tabular}{@{}llcccc@{}}
% \toprule
% \textbf{Model} & \textbf{Method} & \textbf{Avg.} & \textbf{Norm.} & \textbf{Worst} & \textbf{Drop} \\
% \midrule
% \multirow{6}{*}{Llama-3.2-3B} & Model Soup & 29.3 & 57.0 & \textbf{7.2} & 22.1 \\
%  & Task Arithmetic & 37.5 & 72.9 & 25.3 & \textbf{12.2} \\
%  & Dataless L\&S & 35.1 & 68.3 & 10.4 & 24.7 \\
%  & L\&S & 38.3 & 74.6 & 23.7 & 14.6 \\
%  & RegMean & \textbf{41.5} & \textbf{80.7} & 14.2 & 27.3 \\
%  & Fisher Merging & 31.8 & 61.8 & 12.0 & 19.8 \\

% \midrule
% \multirow{6}{*}{Llama-3.1-8B} & Model Soup & 46.6 & 81.1 & 8.3 & 38.3 \\
%  & Task Arithmetic & 48.7 & 84.8 & 31.2 & 17.5 \\
%  & Dataless L\&S & 52.0 & 90.6 & 18.1 & 33.9 \\
%  & L\&S & \textbf{52.7} & \textbf{91.7} & 37.3 & \textbf{15.4} \\
%  & RegMean & 46.3 & 80.6 & 10.9 & 35.4 \\
%  & Fisher Merging & 46.2 & 80.3 & \textbf{5.2} & 41.0 \\
% \bottomrule
% \end{tabular}}
% \caption{MergeBench examples on Llama-3.2-3B and Llama-3.1-8B. Avg. is mean accuracy over five domains. Norm. is normalized performance against source models. Drop is Avg. minus the lowest domain score \citep{he2025mergebenchbenchmark}.}
% \label{tab:avg_score_limit}
% \end{table}

\paragraph{Benchmarks.}
Model fusion benchmarks involve more than a task leaderboard.
They usually define a model pool and a task pool, so methods can be compared under shared source and evaluation settings.
FusionBench \citep{tang2025fusionbenchcomprehensive} gives unified settings for comparing many parameter-level fusion methods across model and task pools.
MergeBench \citep{he2025mergebenchbenchmark} focuses on domain source models and reports retention, generalization, and cost.
Appendix Table~\ref{tab:model_fusion_benchmark_analysis} compares representative resources by modality coverage, model pool, task pool, heterogeneity, fusion type, and evaluation focus.
The comparison shows that current resources still mainly support parameter-level fusion.
Representation-level fusion often relies on drift analysis in method papers.
Behavior-level fusion often borrows task, response, or safety benchmarks from distillation studies \citep{xu2024surveyknowledge,song2026surveyonpolicydistillationlarge}.
Future benchmarks should share source settings and report drift, behavior transfer, judge protocols, and fusion cost.

\definecolor{TakeLevelP}{HTML}{1F77B4}
\definecolor{TakeLevelR}{HTML}{2CA02C}
\definecolor{TakeLevelB}{HTML}{C2410C}
\newcommand{\takelevelbadge}[2][]{%
  \raisebox{0.32ex}[0.9em][0.1em]{%
    \tikz[baseline=(c.base)]%
    \node[circle, fill=TakeLevelP, text=white,
    font=\fontsize{5.5}{5.5}\selectfont\bfseries,
    inner sep=0pt, outer sep=0pt, minimum size=1.0em,
    text height=0.75ex, text depth=0.15ex, draw=none, #1](c){#2};}}
\newcommand{\takeP}{\takelevelbadge[fill=TakeLevelP]{P}}
\newcommand{\takeR}{\takelevelbadge[fill=TakeLevelR]{R}}
\newcommand{\takeB}{\takelevelbadge[fill=TakeLevelB]{B}}

\begin{table}[t]
\centering
\footnotesize
\setlength{\tabcolsep}{3pt}
\renewcommand{\arraystretch}{0.92}
\begin{tabularx}{\linewidth}{@{}>{\raggedright\arraybackslash}Xc>{\raggedleft\arraybackslash}p{0.16\linewidth}@{}}
\toprule
\textbf{Method} & \textbf{Level} & \textbf{Avg.} \\
\midrule
\multicolumn{3}{@{}>{\raggedright\arraybackslash}p{\linewidth}@{}}{\textbf{RLVR Expert Fusion}: Qwen3-4B, 5 domains} \\
\cmidrule{1-3}
TIES & \takeP & 61.00 \\
TA+DARE & \takeP & 60.99 \\
MT-OPD & \takeB & 60.46 \\
\midrule
\multicolumn{3}{@{}>{\raggedright\arraybackslash}p{\linewidth}@{}}{\textbf{Domain Expert Fusion}: Llama-3.1-8B, 5-domain Acc. (\%)} \\
\cmidrule{1-3}
Task Arithmetic & \takeP & 48.7 \\
TIES & \takeP & 46.8 \\
RegMean & \takeR & 46.3 \\
DARE & \takeP & 45.2 \\
\midrule
\multicolumn{3}{@{}>{\raggedright\arraybackslash}p{\linewidth}@{}}{\textbf{Instruct+Code}: AlpacaEval 2.0, HumanEval, MBPP} \\
\cmidrule{1-3}
Task Arithmetic (0 shots) & \takeP & 0.2448 \\
ProDistill (16 shots) & \takeR & 0.2367 \\
ProDistill (32 shots) & \takeR & 0.2380 \\
ProDistill (64 shots) & \takeR & 0.2496 \\
\bottomrule
\end{tabularx}
\vspace{0.8ex}

\begin{tabularx}{\linewidth}{@{}>{\raggedright\arraybackslash}Xrr@{}}
\toprule
\textbf{CLIP setting} & \textbf{ProDistill \takeR} & \textbf{SAMerging \takeB} \\
\midrule
B32/8  & 81.1 & 83.8 \\
B32/14 & 80.5 & 81.2 \\
B32/20 & 77.8 & 77.9 \\
L14/8  & 87.2 & 91.2 \\
L14/14 & 89.0 & 89.1 \\
L14/20 & 86.8 & 85.8 \\
\bottomrule
\end{tabularx}
\vspace{1pt}

{\raggedright\scriptsize CLIP values are average accuracy (\%) under matched 16-shot settings. B32 and L14 denote ViT-B/32 and ViT-L/14; 8, 14, and 20 denote TA-8, TALL-14, and TALL-20.\par}
\caption{Within-paper evidence for Takeaway~2 from M2RL~\citep{wang2026mixmerge}, MergeBench~\citep{he2025mergebenchbenchmark}, ProDistill~\citep{xu2025scalablemodel}, and SAMerging~\citep{dalili2025modelmerging}. Values are reported point estimates, and ProDistill shot counts are shown with the method names. \takeP \ ,\ \takeR \ and \takeB \ denote parameter-, representation-, and behavior-level fusion.}
\label{tab:takeaway_cross_level_results}
\end{table}

\begin{table}[t]
\centering
\footnotesize
\setlength{\tabcolsep}{3pt}
\renewcommand{\arraystretch}{1.08}
\resizebox{\linewidth}{!}{%
\begin{tabular}{@{}lclrr@{}}
\toprule
\textbf{Pair} & \textbf{Levels} & \textbf{Setting / Metric} & \textbf{Single} & \textbf{Hybrid} \\
\midrule
TA / TA+FeatCal & \takeP$\rightarrow$\takeP+\takeR & Llama-3.1-8B, 6-task Avg. & 63.5 & 65.8 \\
KD / Patient KD & \takeB$\rightarrow$\takeB+\takeR & RACE test Acc. (\%) & 58.74 & 60.34 \\
KD / TED & \takeB$\rightarrow$\takeB+\takeR & GLUE dev Avg. & 87.0 & 87.5 \\
w/o Pred / TinyBERT & \takeR$\rightarrow$\takeR+\takeB & 3-task GLUE dev Avg. & 73.5 & 75.6 \\
\bottomrule
\end{tabular}}
\caption{Four within-paper comparisons for Takeaway~3 from FeatCal~\citep{gu2026featcalfeaturecalibrationpostmerging}, Patient KD~\citep{sun2019patientknowledge}, TED~\citep{liang2023lessmore}, and TinyBERT~\citep{jiao2020tinybertdistilling}. Each row compares a single-level baseline with its hybrid counterpart under the same paper setting; scores are not compared across rows.}
\label{tab:takeaway_hybrid_results}
\end{table}

\section{Practical Takeaways}

\noindent\ding{182} \textbf{Fusion methods should be selected under practical constraints.}
Figure~\ref{fig:model-fusion-three-levels-latex} compares their signals, data, and access needs.
The key choice is which signal is available and which failure mode is most likely.
When source models share architecture, initialization, and tokenizer, and their weights are available, parameter-level fusion is often a simple first option.
When drift or internal loss appears, representation-level fusion can use hidden states to find and calibrate layer mismatch.
If only responses are available or models differ, behavior-level fusion is practical.

\noindent\ding{183} {\small\textbf{Cross-level rankings vary by setting and budget.}}
Because models, settings, and budgets differ across papers, Table~\ref{tab:takeaway_cross_level_results} uses only within-paper comparisons.
In M2RL, parameter-level TIES and TA+DARE are close to behavior-level MT-OPD (61.00 and 60.99 versus 60.46), while MT-OPD uses 967.9 GPU-hours of training.
In MergeBench, representation-level RegMean (46.3) lies between parameter-level TIES (46.8) and DARE (45.2).
On Instruct+Code, ProDistill trails Task Arithmetic at 16 and 32 shots but slightly exceeds it at 64 shots (0.2496 versus 0.2448).
Under matched 16-shot CLIP settings, SAMerging has higher reported point estimates than ProDistill in five of six configurations, whereas ProDistill is higher on L14/20.
Thus, cross-level rankings depend on the evaluation setting, data access, and compute budget.

\noindent\ding{184} \textbf{Combining fusion levels can yield a stronger practical pipeline.}
Different levels address complementary failure modes: parameter fusion provides a low-cost target, representation fusion corrects drift or mismatch, and behavior supervision recovers missing outputs when cheaper signals are insufficient.
Across four within-paper comparisons, adding a second fusion level improves the reported score: FeatCal raises Task Arithmetic from 63.5 to 65.8, Patient KD raises RACE test accuracy from 58.74 to 60.34, TED raises the GLUE development average from 87.0 to 87.5, and TinyBERT raises the three-task GLUE development average from 73.5 to 75.6.
Together, the four comparisons span \takeP+\takeR, \takeB+\takeR, and \takeR+\takeB pipelines; they do not imply universal gains.

\section{Challenges and Future Directions}
\label{sec:challenges}

\noindent\ding{182} \textbf{Unclear Theoretical Foundations and Applicability Conditions.}
Model fusion is already used in practice: DeepSeek-V4 independently trains domain experts and consolidates them through on-policy distillation \citep{deepseekai2026deepseekv4}.
Yet fusion quality is usually known only after the target model has been built and evaluated, and no general theory predicts applicability across all three fusion levels.
Source and method selection therefore still rely heavily on trial and error, increasing cost and the risk of capability loss.
Recent theory explains specific cases: \citet{li2026unifiedgeneralization} analyzes parameter averaging under heterogeneous fine-tuning through $L_2$-stability, while other work studies shared initialization, nearby loss basins, or hidden-unit alignment \citep{wortsman2022modelsoups,ainsworth2023gitre,zhou2026demystifyingmergeability}.
These results do not yet provide comparable conditions for representation alignment or behavior distillation.
Theory should link outcomes to source compatibility, fusion data, and target capacity, and identify settings where source capabilities may not be retained.

\noindent\ding{183} \textbf{Difficulty in Aligning Heterogeneous Source Models.}
Source models can differ in architecture, parameterization, tokenizer, or modality interface, leaving no direct correspondence between their components.
This breaks parameter correspondence at the parameter level, makes hidden-state matching harder at the representation level, and complicates behavior distillation when output spaces or reasoning styles differ.
Transport and Merge \citep{cui2026transportmerge} uses optimal transport for cross-architecture LLM fusion, while AdaMMS \citep{du2025adammsmodel} learns coefficients for heterogeneous MLLMs.
Without reliable alignment, source capabilities can interfere rather than combine.
Future work should align architectures and translate representations across model families and domains.

\noindent\ding{184} \textbf{Evaluating Multi-Source Capability Retention.}
Model fusion succeeds only if one target model retains the capabilities of multiple sources.
Aggregate scores can hide source-specific capability loss and interference, making an unsuccessful fusion appear effective.
Appendix Table~\ref{tab:model_fusion_benchmark_analysis} surveys 14 benchmarks and related evaluation resources, but reusable multi-method protocols remain concentrated in a small subset.
FusionBench \citep{tang2025fusionbenchcomprehensive} and MergeBench \citep{he2025mergebenchbenchmark} provide broad evaluation settings, whereas most other resources focus on narrower tasks, individual methods, empirical studies, or search and deployment tooling.
Future benchmarks should use shared source settings and report per-capability retention, worst-task degradation, fusion cost, and compliance with the single-model inference condition.

\noindent\ding{185} \textbf{Preventing Risk Transfer During Fusion.}
Unsafe behavior can survive parameter merging \citep{hammoud2024modelmerging}; LoRATK \citep{liu2024loraas} and Merge Hijacking \citep{yuan2025mergehijacking} further show how malicious updates and backdoors can enter a fused model.
Backdoors can also transfer through behavior distillation \citep{wang2025backweak,demuri2025triggers} and remain in the target model after the source models have been removed at inference time.
Fusion also creates privacy, ownership, and collaboration risks: Merger-as-a-Stealer \citep{lu2025mergeras} studies private information leakage, while Among Us \citep{yang2026amongus} studies malicious contributions in model collaboration.
Because consolidating these attacks may lower the barrier to reproduction or misuse, future work should combine source screening, provenance tracking, contribution attribution, and post-fusion safety testing with defensive mechanisms such as MergeGuard \citep{cong2024haveyou}. Appendix~\ref{app:continual_large_scale_challenges} discusses forgetting and deployment cost.

\section{Conclusion}

We define model fusion through three operations: parameter combination, representation alignment, and behavior distillation. The fused target must run without the complete source models at inference. This boundary supports a three-level taxonomy and exposes level-specific problems in compatibility, capability retention, and risk transfer.

No fusion level dominates across the reviewed settings. Comparisons must account for source compatibility, available signals, fusion data, and compute budgets. Multi-level pipelines can combine an inexpensive initial merge with representation repair or behavior supervision, but the reported gains remain setting dependent. Progress therefore requires applicability conditions for each level, reliable alignment across heterogeneous sources, evaluation of per-source capability retention, and safeguards against risk transfer.

\section*{Limitations}

This survey may miss recent fusion work, especially fast-moving preprints and industrial systems with limited public details.
Relevant papers may also be overlooked because the topic appears as model merging or knowledge transfer.
We collected papers from surveys, benchmarks, and method papers and repeatedly checked their taxonomy and references, but errors may remain.
Some methods combine multiple levels. Our labels identify the main signal used to construct or update the target rather than mutually exclusive pipeline classes, and boundary cases may admit alternative readings.

Our benchmark summary relies on reported results and may not account for model scale, data access, or tuning budget. The cross-paper numbers in Section~4 are illustrative rather than controlled head-to-head comparisons, and some sources report neither variance nor significance. The conclusions are therefore setting-dependent, not universal rankings among fusion levels. Reported fusion costs are likewise not independently normalized across hardware and implementation settings.

\section*{Acknowledgments}

This paper is fully supported by five grants from the Research Grants Council of the Hong Kong Special Administrative Region, China (No.~15215325, 15208824, 15228325, 25208626, T41-517/25-N) and an Innovation and Technology Funds from the Innovation and Technology Commission of the Hong Kong Special Administrative Region, China (No.~ITP/003/26LP).

\bibliography{custom}

\appendix

\newpage
\section{Symbol Definitions}

\noindent
\begin{center}
{\small
\setlength{\tabcolsep}{4pt}
\begin{tabularx}{\linewidth}{@{}p{0.28\linewidth}X@{}}
\toprule
Symbol & Meaning \\
\midrule
\(\mathcal X\) & Input space. \\
\(\mathcal Y\) & Output space. \\
\(x\) & An input instance. \\
\(y\) & An output instance. \\
\(n\) & Number of source models. \\
\(\mathcal S\) & Set of source models, \(\mathcal S=\{M_i^{\mathrm{src}}\}_{i=1}^{n}\). \\
\(M_i^{\mathrm{src}}\) & The \(i\)-th source model. \\
\(p_i^{\mathrm{src}}(y\mid x)\) & Conditional output distribution of the \(i\)-th source model. \\
\(\mathcal T_i\) & Task distribution associated with source \(i\). \\
\(M_\theta^{\mathrm{tgt}}\) & Target model parameterized by \(\theta\). \\
\(\theta\) & Parameters of the target model. \\
\(\theta^\star\) & Fused target parameters in the general retention goal. \\
\(\theta_0\) & Reference parameters for parameter-level fusion. \\
\(\theta_i\) & Parameters of source model \(M_i^{\mathrm{src}}\). \\
\(\Delta_i\) & Source update, \(\Delta_i=\theta_i-\theta_0\). \\
\(\alpha_i\) & Scalar coefficient for source \(i\). \\
\(A_i\) & Operator applied to \(\Delta_i\), such as the identity, masking or rescaling, permutation, or subspace projection. \\
\(p_\theta^{\mathrm{tgt}}(y\mid x)\) & Conditional output distribution of the target model. \\
\(\mathcal L,\ell\) & Set of matched layers and an index in that set, respectively. \\
\(\widetilde r_i^{\ell}(x)\) & Aligned representation of source \(i\) at matched layer \(\ell\). \\
\(r_\theta^\ell(x)\) & Representation of the target model at layer \(\ell\). \\
\(s\) & Input state; for an autoregressive model, \(s=(x,y_{<t})\). \\
\(\mu_{\mathcal D}\) & Empirical state distribution from fixed or target-generated sequences. \\
\(q_{\mathcal S}(\cdot\mid s)\) & Behavior target formed by selecting, aggregating, or aligning source outputs. \\
\(p_\theta^{\mathrm{tgt}}(\cdot\mid s)\) & Target output distribution at state \(s\). \\
\(\mathcal D,\mathcal D_i\) & Optional fusion data and the source-specific fusion-data distribution, respectively; \(\mathcal D\) can be empty. \\
\(\Phi,\Phi_{\mathrm P},\Phi_{\mathrm R},\Phi_{\mathrm B}\) & General fusion mapping and its parameter-, representation-, and behavior-level forms. \\
\(D_{\mathrm{out}}\) & Output-space discrepancy used in the general retention goal. \\
\(D_{\mathrm R}\) & Representation discrepancy. \\
\(D_{\mathrm B}\) & Behavior-distribution discrepancy. \\
\bottomrule
\end{tabularx}
\captionof{table}{Notation used in the model fusion formulation.}
\label{tab:notation}
}
\end{center}

\section{Applications}
\label{app:applications}

We cover continual learning, capability integration, safety, and compression.

\paragraph{Model Fusion in Continual Learning.}
In continual learning, model fusion can add new task or domain knowledge while limiting forgetting.
AIMMerging \citep{feng2025aimmergingadaptive} and NUFILT \citep{qiu2025nullspace} apply parameter-level fusion to add new task updates while reducing forgetting and interference.
K-Merge \citep{shenaj2025kmerge} extends this setting to online LoRA fusion for on-device LLMs.
RECALL \citep{wang2025recallrepresentation} uses hidden representations for hierarchical fusion without historical data.
SDFT \citep{shenfeld2026selfdistillation} uses behavior-level fusion to learn new skills while reducing forgetting.

\paragraph{Multi-Task Learning and Domain Capability Integration.}
Model fusion integrates source models trained for different tasks, domains, or languages.
Compared with training one model on mixed task data, it can reuse existing source models and reduce reliance on original data or full retraining \citep{jin2023datalessknowledge,yang2024adamerging}.
Language Specific Model Merging \citep{dmonte2026improvingtraining} fuses language-specific models to lower multilingual training and update costs.
SurgeryV2 \citep{yang2024surgeryv2bridging} and FeatCal \citep{gu2026featcalfeaturecalibrationpostmerging} repair representation drift after fusion.
FuseLLM \citep{wan2024knowledgefusion} and DeepSeek-V4 \citep{deepseekai2026deepseekv4} use behavior-level fusion to integrate source capabilities.

\paragraph{Safety and Control.}
For safety control, model fusion can transfer, keep, or weaken behavior attributes after training.
SafeMERGE \citep{djuhera2025safemergepreserving} and Fuse to Forget \citep{zaman2024fuseforget} use parameter-level fusion to preserve safety or reduce unwanted behavior.
Safety Realignment \citep{yi2024safetyrealignment} uses subspace-oriented model fusion to realign unsafe models.
Multilingual Safety Alignment via Self-Distillation \citep{2605.02971} transfers safety behavior across languages through behavior-level fusion.
However, unsafe source models can also propagate misalignment during fusion \citep{hammoud2024modelmerging}.

\paragraph{Model Compression.}
Model compression uses source models to build smaller target models with similar capabilities.
LoRA soups \citep{prabhakar2025lorasoups} and LoRM \citep{salami2025closedform} can fold several lightweight modules into one target module.
DeepSeek-R1 \citep{ai2025deepseekr1} transfers reasoning patterns into six dense models with 1.5B to 70B parameters.
Nemotron-Cascade 2 \citep{yang2026nemotroncascade} builds a compact 30B MoE model, with 3B active parameters, for math, code, and agentic tasks.
Smaller target models can lower serving cost and speed up inference in resource-limited settings.

\section{Parameter-Level Fusion Analysis}

This appendix compares representative parameter-level fusion methods by source-model relation, fusion object, and evaluated backbones or settings.

\definecolor{PFPlaceholder}{HTML}{6B7280}
\definecolor{PFGroup}{HTML}{DAE8FC}
\definecolor{PFSourceSame}{HTML}{2563EB}
\definecolor{PFSourceCkpt}{HTML}{A16207}
\definecolor{PFObjWeight}{HTML}{1F77B4}
\definecolor{PFObjTask}{HTML}{2CA02C}
\definecolor{PFObjAgg}{HTML}{9467BD}
\definecolor{PFObjPeft}{HTML}{C2410C}
\definecolor{PFArchCV}{HTML}{0E7490}
\definecolor{PFArchLM}{HTML}{2563EB}
\definecolor{PFArchED}{HTML}{7E22CE}
\definecolor{PFArchLLM}{HTML}{BE185D}
\definecolor{PFArchDiff}{HTML}{A16207}
\newcommand{\pfbadge}[2][]{%
  \tikz[baseline=(c.base)]%
  \node[circle, fill=PFPlaceholder, text=white, font=\scriptsize\bfseries,
  inner sep=0pt, minimum size=2.05em, draw=none, #1](c){\strut #2};}
\newcommand{\pfobjbadge}[2][]{%
  \tikz[baseline=(c.base)]%
  \node[circle, fill=PFPlaceholder, text=white, font=\scriptsize\bfseries,
  inner sep=0pt, minimum size=2.05em, draw=none, #1](c){\strut #2};}
\newcommand{\pfplaceholder}{\pfbadge{--}\ placeholder}
\newcommand{\pfsrcS}{\pfbadge[fill=PFSourceSame]{S}}
\newcommand{\pfsrcC}{\pfbadge[fill=PFSourceCkpt]{C}}
\newcommand{\pfobjW}{\pfobjbadge[fill=PFObjWeight]{W}}
\newcommand{\pfobjT}{\pfobjbadge[fill=PFObjTask]{T}}
\newcommand{\pfobjA}{\pfobjbadge[fill=PFObjAgg]{A}}
\newcommand{\pfobjP}{\pfobjbadge[fill=PFObjPeft]{P}}
\newcommand{\pfarchCV}{\pfobjbadge[fill=PFArchCV]{CV}}
\newcommand{\pfarchLM}{\pfobjbadge[fill=PFArchLM]{LM}}
\newcommand{\pfarchED}{\pfobjbadge[fill=PFArchED]{ED}}
\newcommand{\pfarchLLM}{\pfobjbadge[fill=PFArchLLM]{LLM}}
\newcommand{\pfarchDiff}{\pfobjbadge[fill=PFArchDiff]{DIF}}

\begin{table*}[p]
\centering
\scriptsize
\setlength{\tabcolsep}{1.8pt}
\renewcommand{\arraystretch}{1.05}
\begin{tabular*}{\textwidth}{@{\extracolsep{\fill}}lllll@{}}
\toprule
\textbf{Method} & \textbf{Venue} & \textbf{Source relation} & \textbf{Fusion object} & \textbf{Evaluated backbones/settings} \\
\midrule
\multicolumn{5}{@{}l}{\textit{Arithmetic rules}} \\
\midrule
SWA \citeyearpar{izmailov2018averagingweights} & UAI'18 & \pfsrcC\ trajectory & \pfobjW\ checkpoints & \pfarchCV\ CNNs \\
MWA \citeyearpar{yu2025parameterefficient} & arXiv'25 & \pfsrcC\ one-run ckpts. & \pfobjP\ LoRA ckpts. & \pfarchLLM\ Gemma-2B \\
Model Soups \citeyearpar{wortsman2022modelsoups} & ICML'22 & \pfsrcS\ same-base FT & \pfobjW\ weights & \pfarchCV\ CLIP/ViT \\
Fisher \citeyearpar{matena2022mergingmodels} & NeurIPS'22 & \pfsrcS\ same-init. & \pfobjW\ weights & \pfarchCV\ \pfarchLM\ ViT/BERT \\
Task Arith. \citeyearpar{ilharco2023editing} & ICLR'23 & \pfsrcS\ same-base & \pfobjT\ task vectors & \pfarchCV\ \pfarchLLM\ \pfarchED\ CLIP/GPT-2/T5 \\
TIES \citeyearpar{yadav2023tiesmerging} & NeurIPS'23 & \pfsrcS\ same-base & \pfobjT\ deltas & \pfarchCV\ \pfarchED\ ViT/T5 \\
DARE \citeyearpar{yu2024languagemodels} & ICML'24 & \pfsrcS\ same-base & \pfobjT\ deltas & \pfarchLM\ \pfarchLLM\ BERT/Llama \\
DELLA \citeyearpar{deep2024dellamerging} & arXiv'24 & \pfsrcS\ same-base & \pfobjT\ deltas & \pfarchLLM\ Llama-2 \\
Diff. Soup \citeyearpar{biggs2024diffusionsoup} & ECCV'24 & \pfsrcS\ same-base & \pfobjW\ weights & \pfarchDiff\ T2I diffusion \\
\midrule
\multicolumn{5}{@{}l}{\textit{Subspace-based methods}} \\
\midrule
KnOTS \citeyearpar{stoica2024modelmerging} & ICLR'25 & \pfsrcS\ shared-base LoRA & \pfobjP\ \pfobjT\ LoRA updates & \pfarchCV\ \pfarchLLM\ CLIP/Llama3 \\
DOP \citeyearpar{yang2026continual} & NeurIPS'25 & \pfsrcS\ sequential & \pfobjT\ \pfobjA\ task/merged deltas & \pfarchCV\ \pfarchED\ ViT/Flan-T5 \\
Iso-C/CTS \citeyearpar{marczak2025notask} & ICML'25 & \pfsrcS\ same-base & \pfobjA\ aggregate & \pfarchCV\ \pfarchLLM\ CLIP/LLMs \\
TSV \citeyearpar{gargiulo2025tasksingular} & CVPR'25 & \pfsrcS\ same-base & \pfobjT\ task matrices & \pfarchCV\ CLIP-ViT \\
STAR \citeyearpar{lee2025star} & NAACL'25 & \pfsrcS\ same-base & \pfobjT\ spectral deltas & \pfarchED\ Flan-T5 \\
DC-Merge \citeyearpar{zhang2026dcmerge} & CVPR'26 & \pfsrcS\ same-base & \pfobjT\ \pfobjP\ full/LoRA deltas & \pfarchCV\ \pfarchLLM\ CV/VLM \\
Orthogonal \citeyearpar{yang2026orthogonal} & ICML'26 & \pfsrcS\ same-base & \pfobjT\ \pfobjP\ orthogonal deltas & \pfarchLLM\ LLM/VLM \\
Extra-Merge \citeyearpar{zhou2026extramerge} & ICML'26 & \pfsrcC\ pretrain ckpts. & \pfobjW\ checkpoints & \pfarchLLM\ GPT-2/Pythia/LLaMA \\
ResMerge \citeyearpar{sun2026resmerge} & arXiv'26 & \pfsrcS\ RL experts & \pfobjT\ spectral residuals & \pfarchLLM\ reasoning LLMs \\
\bottomrule
\end{tabular*}
\caption{Parameter-level fusion methods compared by source-model relation, fusion object, and evaluated backbones or settings. Method groups follow the taxonomy in Section~3 and Figure~\ref{fig:model-fusion-overview}.}
\label{tab:param_level_fusion_analysis}
\end{table*}

\begin{table*}[p]
\ContinuedFloat
\centering
\scriptsize
\setlength{\tabcolsep}{1.8pt}
\renewcommand{\arraystretch}{1.05}
\begin{tabular*}{\textwidth}{@{\extracolsep{\fill}}lllll@{}}
\toprule
\textbf{Method} & \textbf{Venue} & \textbf{Source relation} & \textbf{Fusion object} & \textbf{Evaluated backbones/settings} \\
\midrule
\multicolumn{5}{@{}l}{\textit{Optimization-based methods}} \\
\midrule
AdaMerging \citeyearpar{yang2024adamerging} & ICLR'24 & \pfsrcS\ same-base & \pfobjT\ vectors + coeffs. & \pfarchCV\ CLIP-ViT \\
AWD \citeyearpar{xiong2024multitask} & arXiv'24 & \pfsrcS\ same-base & \pfobjT\ disentangled & \pfarchCV\ \pfarchLM\ ViT/RoBERTa \\
WUDI \citeyearpar{cheng2025whoever} & ICML'25 & \pfsrcS\ same-base & \pfobjA\ merged vector & \pfarchCV\ \pfarchLM\ \pfarchLLM\ ViT/RoBERTa/Llama \\
DOGE \citeyearpar{wei2025modelingmulti} & ICML'25 & \pfsrcS\ same-base & \pfobjT\ \pfobjA\ merged update & \pfarchCV\ \pfarchLM\ \pfarchLLM\ vision/NLP \\
GCWM \citeyearpar{wang2026geometry} & arXiv'26 & \pfsrcC\ continual & \pfobjA\ cumulative update & \pfarchLLM\ Qwen3 \\
MergOPT \citeyearpar{yang2026mergopt} & ICLR'26 & \pfsrcS\ same-base & \pfobjW\ optimized weights & \pfarchCV\ \pfarchLLM\ vision/LLMs \\
EvoGM \citeyearpar{jiang2026evogm} & ICML'26 & \pfsrcS\ same-base & \pfobjT\ vectors + coeffs. & \pfarchLLM\ Qwen/Llama \\
ODE-M \citeyearpar{lin2026odemerging} & ICML'26 & \pfsrcC\ continual & \pfobjA\ merged trajectory & \pfarchCV\ CLIP-ViT \\
SWUDI \citeyearpar{wei2026swudi} & arXiv'26 & \pfsrcS\ same-base & \pfobjT\ \pfobjA\ spectral update & \pfarchLLM\ LLM/multimodal \\
\midrule
\multicolumn{5}{@{}l}{\textit{Module merging}} \\
\midrule
AdapterSoup \citeyearpar{chronopoulou2023adaptersoupweight} & EACL-F'23 & \pfsrcS\ shared-base & \pfobjP\ adapters & \pfarchLLM\ GPT-2 \\
HydraOpt \citeyearpar{ceritli-etal-2025-hydraopt} & EMNLP'25 & \pfsrcS\ shared-base & \pfobjP\ adapters & \pfarchLLM\ Llama/Qwen \\
LoRASoup \citeyearpar{prabhakar2025lorasoups} & COLING-I'25 & \pfsrcS\ shared-base & \pfobjP\ LoRAs & \pfarchLLM\ Llama-7B \\
Lora-Flow \citeyearpar{wang2024loraflow} & ACL'24 & \pfsrcS\ shared-base & \pfobjP\ LoRAs & \pfarchLLM\ Llama-2 \\
RobustMerge \citeyearpar{zeng2025robustmerge} & NeurIPS'25 & \pfsrcS\ shared-base & \pfobjP\ low-rank & \pfarchLLM\ MLLMs \\
Adaptive LoRA \citeyearpar{miyano2025adaptiveloramerge} & ACL-F'25 & \pfsrcS\ shared-base & \pfobjP\ pruned LoRA & \pfarchLLM\ low-resource gen. \\
NSC \citeyearpar{lee2026labelfreelora} & CVPR'26 & \pfsrcS\ shared-base & \pfobjP\ compressed LoRAs & \pfarchCV\ \pfarchLLM\ CV/NLI/VLM \\
CtM \citeyearpar{he2026compressthenmerge} & ICML'26 & \pfsrcS\ shared-base & \pfobjP\ single LoRA & \pfarchLLM\ Llama-3 \\
PDA/GAM \citeyearpar{sun2026preferencedelta} & arXiv'26 & \pfsrcS\ preference-tuned & \pfobjP\ aligned deltas & \pfarchLLM\ Qwen/Tulu \\
\bottomrule
\end{tabular*}
\vspace{0.35ex}
\begin{tabular*}{\textwidth}{@{\extracolsep{\fill}}>{\bfseries}ll@{}}
Source & \pfsrcS\ same-base coordinates; \pfsrcC\ checkpoint/continual trajectory.\\
Object & \pfobjW\ full weights; \pfobjT\ per-task update; \pfobjA\ merged update; \pfobjP\ PEFT/LoRA/adapter.\\
Backbone & \pfarchCV\ vision; \pfarchLM\ encoder LM; \pfarchED\ encoder--decoder; \pfarchLLM\ decoder LLM; \pfarchDiff\ diffusion.\\
Note & Badges are non-exclusive; grouping follows Figure~\ref{fig:model-fusion-overview}.\\
\end{tabular*}
\caption[]{Parameter-level fusion methods compared by source-model relation, fusion object, and evaluated backbones or settings (continued).}
\end{table*}

\section{Representation-Level Fusion Analysis}

This appendix compares representation methods by source relation, fused scope, and evaluation setting.
\definecolor{FSFull}{HTML}{1F77B4}
\definecolor{FSLinear}{HTML}{2CA02C}
\definecolor{FSAttn}{HTML}{9467BD}
\definecolor{FSNorm}{HTML}{D97706}
\definecolor{FSAdapter}{HTML}{C2410C}
\definecolor{FSBackboneEnc}{HTML}{0E7490}
\definecolor{FSBackboneDec}{HTML}{2563EB}
\definecolor{FSBackboneED}{HTML}{7E22CE}
\definecolor{FSBackboneMM}{HTML}{BE185D}
\definecolor{FSBackboneCkpt}{HTML}{A16207}
\definecolor{FSSourceSame}{HTML}{2563EB}
\definecolor{FSSourceAligned}{HTML}{059669}
\definecolor{FSSourceHetero}{HTML}{D97706}
\definecolor{FSSourceTeacher}{HTML}{7C3AED}
\newcommand{\fsbadge}[2][]{%
  \tikz[baseline=(c.base)]%
  \node[circle, fill=FSFull, text=white, font=\scriptsize\bfseries,
  inner sep=0.85pt, minimum size=1.35em, draw=none, #1](c){\strut #2};}
\newcommand{\srcS}{\fsbadge[fill=FSSourceSame]{S}}
\newcommand{\srcA}{\fsbadge[fill=FSSourceAligned]{A}}
\newcommand{\srcH}{\fsbadge[fill=FSSourceHetero]{H}}
\newcommand{\srcT}{\fsbadge[fill=FSSourceTeacher]{T}}
\newcommand{\paramF}{\fsbadge[fill=FSFull]{F}}
\newcommand{\paramL}{\fsbadge[fill=FSLinear]{L}}
\newcommand{\paramA}{\fsbadge[fill=FSAttn]{A}}
\newcommand{\paramN}{\fsbadge[fill=FSNorm]{N}}
\newcommand{\paramP}{\fsbadge[fill=FSAdapter]{P}}
\newcommand{\archEnc}{\fsbadge[fill=FSBackboneEnc]{E}}
\newcommand{\archDec}{\fsbadge[fill=FSBackboneDec]{D}}
\newcommand{\archED}{\fsbadge[fill=FSBackboneED]{T}}
\newcommand{\archMM}{\fsbadge[fill=FSBackboneMM]{M}}
\newcommand{\archCkpt}{\fsbadge[fill=FSBackboneCkpt]{C}}
\begin{table*}[p]
\centering
\scriptsize
\setlength{\tabcolsep}{2.6pt}
\renewcommand{\arraystretch}{1.08}
\begin{tabularx}{\textwidth}{@{}>{\raggedright\arraybackslash}p{0.235\textwidth}>{\raggedright\arraybackslash}p{0.095\textwidth}>{\raggedright\arraybackslash}p{0.175\textwidth}>{\raggedright\arraybackslash}p{0.155\textwidth}X@{}}
\toprule
\textbf{Method} & \textbf{Venue} & \textbf{Source relation} & \textbf{Fused params/modules} & \textbf{Evaluated backbones/settings} \\
\midrule
\multicolumn{5}{@{}l}{\textit{Weighting and representation matching}} \\
\midrule
\textsuperscript{\dag}REPAIR \citep{jordan2023repair} & ICLR'23 & \srcA\ same architecture after permutation alignment & \paramN & \archEnc\ CNN classifiers \\
ZipIt! \citep{stoica2024zipitmerging} & ICLR'24 & \srcA\ architecture-compatible sources & \paramL\ \paramA & \archEnc\ vision Transformers/classifiers \\
Transformer Fusion \citep{imfeld2024transformerfusion} & ICLR'24 & \srcA\ aligned Transformer variants & \paramL\ \paramA & \archEnc\ ViT and BERT encoders \\
AIM \citep{nobari2025activationinformed} & NeurIPS'25 & \srcS\ same-base LLM checkpoints & \paramF & \archDec\ decoder-only LLMs \\
ACM \citep{yao2025activationguided} & NeurIPS'25 & \srcS\ same-base LLM checkpoints & \paramF & \archDec\ decoder-only LLMs \\
MAGIC \citep{li2025magicachieving} & arXiv'25 & \srcS\ same-base or aligned sources & \paramF\ \paramN & \archEnc\ \archDec\ CV and Llama merging \\
Merging Beyond \citep{yao2026mergingbeyond} & arXiv'26 & \srcS\ sequential same-backbone updates & \paramF\ \paramL & \archDec\ streaming LLM updates \\
\midrule
\multicolumn{5}{@{}l}{\textit{Closed-form representation solvers}} \\
\midrule
RegMean \citep{jin2023datalessknowledge} & ICLR'23 & \srcS\ same architecture and tokenizer & \paramL & \archEnc\ \archED\ RoBERTa/DeBERTa and T5 \\
RegMean++ \citep{nguyen2025regmeanenhancing} & arXiv'25 & \srcA\ same-family compatible models & \paramL & \archEnc\ \archED\ \archDec\ encoder, enc--dec, decoder-only \\
LoRM \citep{salami2025closedform} & ICLR'25 & \srcS\ PEFT modules over compatible bases & \paramL\ \paramP & \archEnc\ \archED\ ViT-B/16 and T5-small \\
IterIS \citep{chen2025iterisiterative} & CVPR'25 & \srcS\ compatible LoRA adapters & \paramP & \archMM\ text-to-image, VLM, and LLM adapters \\
LOT-Merging \citep{sun2025towardsminimizing} & NeurIPS'25 & \srcS\ same-base task-vector checkpoints & \paramL\ \paramN & \archEnc\ ViT and RoBERTa checkpoints \\
FeatCal \citep{gu2026featcalfeaturecalibrationpostmerging} \takeP+\takeR & arXiv'26 & \srcS\ same-base task experts and merged models & \paramL\ \paramN & \archEnc\ \archED\ \archDec\ CLIP, FLAN-T5, and Llama \\
\midrule
\multicolumn{5}{@{}l}{\textit{Backpropagation-based representation transfer}} \\
\midrule
Patient KD \citep{sun2019patientknowledge} \takeB+\takeR & EMNLP'19 & \srcT\ teacher--student fusion & \paramF & \archEnc\ BERT-style encoders \\
TinyBERT \citep{jiao2020tinybertdistilling} \takeR+\takeB & EMNLP Findings'20 & \srcT\ teacher--student fusion & \paramF & \archEnc\ BERT-style encoders \\
TED \citep{liang2023lessmore} \takeB+\takeR & ICML'23 & \srcT\ teacher--student fusion & \paramF\ \paramP & \archEnc\ language-model compression \\
Representation Surgery \citep{yang2024representationsurgery} \takeP+\takeR & ICML'24 & \srcS\ same-base multi-task merged models & \paramP & \archEnc\ encoder-based multi-task models \\
Surgeryv2 \citep{yang2024surgeryv2bridging} \takeP+\takeR & arXiv'24 & \srcS\ same-base multi-task merged models & \paramP & \archEnc\ ViT and BERT multi-task models \\
ProbSurgery \citep{wei2025representationsurgery} \takeP+\takeR & ICML'25 & \srcS\ same-base multi-task merged models & \paramP & \archEnc\ multi-task model merging \\
RECALL \citep{wang2025recallrepresentation} & EMNLP'25 & \srcS\ continual same-family checkpoints & \paramL\ \paramN & \archCkpt\ in-domain checkpoint sequences \\
NUFILT \citep{qiu2025nullspace} & ICLR'26 & \srcS\ continual same-backbone checkpoints & \paramF\ \paramP & \archCkpt\ data-free continual merging \\
OPRD \citep{yang2026oprd} & arXiv'26 & \srcT\ \srcH\ cross-family teacher & \paramF\ \paramP & \archDec\ heterogeneous reasoning LLMs \\
\bottomrule
\end{tabularx}
\vspace{0.35ex}
\begin{tabularx}{\textwidth}{@{}>{\bfseries}p{0.16\textwidth}X@{}}
Source relation &
\srcS\ same base, tokenizer, or checkpoint trajectory;\quad
\srcA\ architecture-compatible sources requiring alignment/matching;\quad
\srcH\ heterogeneous or projection-needed sources;\quad
\srcT\ teacher--student representation fusion.\\
Fused scope &
\paramF\ full parameters or task vectors;\quad
\paramL\ linear/projection weights;\quad
\paramA\ attention components;\quad
\paramN\ normalization, bias, or activation statistics;\quad
\paramP\ projection, LoRA, adapter, or repair module.\\
Backbones/settings &
\archEnc\ encoder or vision backbone;\quad
\archDec\ decoder-only LLM;\quad
\archED\ encoder--decoder;\quad
\archMM\ multimodal or vision--language setting;\quad
\archCkpt\ checkpoint sequence or continual-merging setting.\\
Cross-level labels &
\takeP+\takeR\ parameter fusion with representation calibration or repair;\quad
\takeB+\takeR\ and \takeR+\takeB\ output distillation combined with representation matching.\\
Boundary cases &
\textsuperscript{\dag}Non-LLM representation repair included because it motivates representation-level post-merge correction; teacher--student methods are treated as representation-level fusion when intermediate hidden states or attention maps provide the main fusion signal.
\end{tabularx}
\caption{Representation-level fusion methods compared by source-model relation, fused parameter/module scope, and evaluated model backbones or settings. Method groups follow the taxonomy in Section~3.}
\label{tab:repr_level_fusion_analysis}
\end{table*}

\section{Behavior-Level Fusion Analysis}

This appendix compares behavior methods by source relation, signal, and evaluation setting.

\definecolor{BFPlaceholder}{HTML}{6B7280}
\definecolor{BFSourceTeacher}{HTML}{7C3AED}
\definecolor{BFSourceMulti}{HTML}{2563EB}
\definecolor{BFSourceBlack}{HTML}{D97706}
\definecolor{BFSourceSelf}{HTML}{059669}
\definecolor{BFSourceHetero}{HTML}{BE185D}
\definecolor{BFSignalDist}{HTML}{1F77B4}
\definecolor{BFSignalDemo}{HTML}{2CA02C}
\definecolor{BFSignalFeed}{HTML}{C2410C}
\definecolor{BFSignalTraj}{HTML}{9467BD}
\definecolor{BFStateOff}{HTML}{64748B}
\definecolor{BFStateOn}{HTML}{DC2626}
\definecolor{BFStateSemi}{HTML}{A16207}
\definecolor{BFArchEnc}{HTML}{0E7490}
\definecolor{BFArchDec}{HTML}{2563EB}
\definecolor{BFArchChat}{HTML}{BE185D}
\definecolor{BFArchMM}{HTML}{7E22CE}
\definecolor{BFArchAgent}{HTML}{A16207}

\newcommand{\bfbadge}[2][]{%
  \tikz[baseline=(c.base)]%
  \node[circle, fill=BFPlaceholder, text=white, font=\scriptsize\bfseries,
  inner sep=0.85pt, minimum size=1.35em, draw=none, #1](c){\strut #2};}
\newcommand{\bfsrcT}{\bfbadge[fill=BFSourceTeacher]{T}}
\newcommand{\bfsrcM}{\bfbadge[fill=BFSourceMulti]{M}}
\newcommand{\bfsrcB}{\bfbadge[fill=BFSourceBlack]{B}}
\newcommand{\bfsrcS}{\bfbadge[fill=BFSourceSelf]{S}}
\newcommand{\bfsrcH}{\bfbadge[fill=BFSourceHetero]{H}}
\newcommand{\bfsigD}{\bfbadge[fill=BFSignalDist]{D}}
\newcommand{\bfsigX}{\bfbadge[fill=BFSignalDemo]{X}}
\newcommand{\bfsigF}{\bfbadge[fill=BFSignalFeed]{F}}
\newcommand{\bfsigR}{\bfbadge[fill=BFSignalTraj]{R}}
\newcommand{\bfstOff}{\bfbadge[fill=BFStateOff]{O}}
\newcommand{\bfstOn}{\bfbadge[fill=BFStateOn]{P}}
\newcommand{\bfstSemi}{\bfbadge[fill=BFStateSemi]{S}}
\newcommand{\bfarchEnc}{\bfbadge[fill=BFArchEnc]{E}}
\newcommand{\bfarchDec}{\bfbadge[fill=BFArchDec]{D}}
\newcommand{\bfarchChat}{\bfbadge[fill=BFArchChat]{C}}
\newcommand{\bfarchMM}{\bfbadge[fill=BFArchMM]{M}}
\newcommand{\bfarchAgent}{\bfbadge[fill=BFArchAgent]{A}}

\begin{table*}[p]
\centering
\scriptsize
\setlength{\tabcolsep}{2.4pt}
\renewcommand{\arraystretch}{1.05}
\begin{tabularx}{\textwidth}{@{}>{\raggedright\arraybackslash}p{0.245\textwidth}>{\raggedright\arraybackslash}p{0.075\textwidth}>{\raggedright\arraybackslash}p{0.16\textwidth}>{\raggedright\arraybackslash}p{0.225\textwidth}X@{}}
\toprule
\textbf{Method} & \textbf{Venue} & \textbf{Source relation} & \textbf{Behavior signal} & \textbf{Evaluated backbones/settings} \\
\midrule
\multicolumn{5}{@{}l}{\textit{Distribution fusion}} \\
\midrule
Knowledge Distillation \citep{hinton2015distillingknowledge} & NeurIPS'15 & \bfsrcT\ teacher--student & \bfsigD\ \bfstOff\ soft outputs & general neural networks \\
DistilBERT \citep{sanh2019distilbertdistilled} & NeurIPS'19 & \bfsrcT\ BERT teacher--student & \bfsigD\ \bfstOff\ token distributions & \bfarchEnc\ BERT encoders \\
Patient KD \citep{sun2019patientknowledge} \takeB+\takeR & EMNLP'19 & \bfsrcT\ BERT teacher--student & \bfsigD\ \bfstOff\ outputs + hidden states & \bfarchEnc\ BERT encoders \\
TinyBERT \citep{jiao2020tinybertdistilling} \takeR+\takeB & EMNLP Findings'20 & \bfsrcT\ BERT teacher--student & \bfsigD\ \bfstOff\ logits + representations & \bfarchEnc\ BERT encoders \\
TED \citep{liang2023lessmore} \takeB+\takeR & ICML'23 & \bfsrcT\ LM teacher--student & \bfsigD\ \bfstOff\ outputs + filtered reps. & \bfarchEnc\ encoder LMs \\
InfiGFusion \citep{wang2026infigfusion} & NeurIPS'25 & \bfsrcM\ multi-source LLMs & \bfsigD\ \bfstOff\ logit geometry & \bfarchDec\ decoder-only LLMs \\
\midrule
\multicolumn{5}{@{}l}{\textit{Demonstration fusion}} \\
\midrule
Instruction Distillation \citep{sun2023instructiondistillation} & arXiv'23 & \bfsrcT\ stronger teacher & \bfsigX\ \bfstOff\ instruction responses & \bfarchDec\ LLM rankers \\
GPT4All \citep{anand2023gpt4alltraining} & GitHub'23 & \bfsrcB\ API teacher & \bfsigX\ \bfstOff\ assistant demos & \bfarchChat\ chatbot tuning \\
Distilling Step-by-Step \citep{hsieh2023distillingstep} & ACL'23 & \bfsrcT\ larger LLM teacher & \bfsigX\ \bfstOff\ rationales and labels & \bfarchDec\ small reasoning LMs \\
Teaching Small LMs to Reason \citep{magister2023teachingsmall} & ACL'23 & \bfsrcT\ reasoning teacher & \bfsigX\ \bfstOff\ CoT rationales & \bfarchDec\ small LMs \\
\midrule
\multicolumn{5}{@{}l}{\textit{Feedback fusion}} \\
\midrule
FuseChat \citep{wan2024fusechatknowledge} & EMNLP'25 & \bfsrcM\ chat-model sources & \bfsigX\ \bfsigF\ \bfstOff\ responses and preferences & \bfarchChat\ chat fusion \\
Zephyr \citep{tunstall2023zephyrdirect} & COLM'24 & \bfsrcT\ aligned teacher & \bfsigF\ \bfstOff\ preference data & \bfarchChat\ chat alignment \\
InfiFPO \citep{gu2026infifpo} & NeurIPS'25 & \bfsrcM\ source preferences & \bfsigF\ \bfstOff\ preference optimization & \bfarchDec\ LLM fusion \\
\bottomrule
\end{tabularx}
\caption{Behavior-level fusion methods compared by source-model relation, behavior signal, state distribution, and evaluated backbones or settings. Method groups follow the taxonomy in Section~3.3.}
\label{tab:behavior_level_fusion_analysis}
\end{table*}

\begin{table*}[p]
\ContinuedFloat
\centering
\scriptsize
\setlength{\tabcolsep}{2.4pt}
\renewcommand{\arraystretch}{1.05}
\begin{tabularx}{\textwidth}{@{}>{\raggedright\arraybackslash}p{0.245\textwidth}>{\raggedright\arraybackslash}p{0.075\textwidth}>{\raggedright\arraybackslash}p{0.16\textwidth}>{\raggedright\arraybackslash}p{0.225\textwidth}X@{}}
\toprule
\textbf{Method} & \textbf{Venue} & \textbf{Source relation} & \textbf{Behavior signal} & \textbf{Evaluated backbones/settings} \\
\midrule
\multicolumn{5}{@{}l}{\textit{On-policy and trajectory-level fusion}} \\
\midrule
\textsuperscript{\dag}DAgger \citep{ross2011reductionimitation} & AISTATS'11 & \bfsrcT\ expert policy & \bfsigR\ \bfstOn\ learner-state actions & \bfarchAgent\ imitation learning \\
GKD / OPD \citep{agarwal2024policydistillation} & ICLR'24 & \bfsrcT\ teacher on student states & \bfsigD\ \bfsigR\ \bfstOn\ self-generated sequences & \bfarchDec\ autoregressive LMs \\
Self-Distilled Reasoner \citep{zhao2026selfdistilled} & arXiv'26 & \bfsrcS\ self-distillation & \bfsigX\ \bfsigR\ \bfstOn\ reasoning traces & \bfarchDec\ reasoning LLMs \\
SODA \citep{chen2026soda} & arXiv'26 & \bfsrcB\ black-box teacher & \bfsigX\ \bfsigR\ \bfstSemi\ semi-on-policy data & \bfarchDec\ black-box distillation \\
Lightning OPD \citep{wu2026lightning} & arXiv'26 & \bfsrcT\ teacher-logit source & \bfsigD\ \bfstSemi\ offline OPD signals & \bfarchDec\ reasoning LLMs \\
TIP \citep{xu2026tip} & arXiv'26 & \bfsrcT\ sampled-token teacher & \bfsigD\ \bfsigR\ \bfstOn\ token-importance signals & \bfarchDec\ token-efficient OPD \\
Demystifying OPD \citep{luo2026demystifying} & arXiv'26 & \bfsrcT\ rollout teacher & \bfsigD\ \bfsigR\ \bfstOn\ stabilized token signals & \bfarchDec\ OPD stabilization \\
Video-OPD \citep{li2026videoopd} & arXiv'26 & \bfsrcH\ multimodal teacher & \bfsigR\ \bfstOn\ video trajectories & \bfarchMM\ video grounding MLLMs \\
X-OPD \citep{cao2026xopd} & arXiv'26 & \bfsrcH\ cross-modal teacher & \bfsigR\ \bfstOn\ speech trajectories & \bfarchMM\ speech LLM alignment \\
CORD \citep{jing2026cord} & arXiv'26 & \bfsrcS\ internal text teacher & \bfsigD\ \bfsigF\ \bfsigR\ \bfstOn\ token KL + reward & \bfarchMM\ audio LLM reasoning \\
Entropy-Aware OPD \citep{jin2026entropyaware} & ICML'26 & \bfsrcT\ uncertainty-aware teacher & \bfsigD\ \bfsigR\ \bfstOn\ entropy-adaptive KL & \bfarchDec\ reasoning LLMs \\
SCOPE \citep{zheng2026scope} & arXiv'26 & \bfsrcT\ correctness-routed teacher & \bfsigD\ \bfsigR\ \bfstOn\ calibrated KL/MLE & \bfarchDec\ reasoning LLMs \\
TCOD \citep{wang2026tcod} & arXiv'26 & \bfsrcT\ multi-turn-state teacher & \bfsigD\ \bfsigR\ \bfstOn\ temporal curriculum & \bfarchAgent\ multi-turn LLM agents \\
SOD \citep{zhong2026sod} & arXiv'26 & \bfsrcT\ tool-state teacher & \bfsigD\ \bfsigR\ \bfstOn\ step-weighted tokens & \bfarchAgent\ small tool-use agents \\
SFD / Lookahead Group Reward \citep{liu2026supervisionfidelity} & arXiv'26 & \bfsrcT\ long-prefix teacher & \bfsigD\ \bfsigF\ \bfsigR\ \bfstOn\ lookahead rewards & \bfarchDec\ long-horizon reasoning \\
POPD / TOPD \citep{zhang2026fullrollouts} & arXiv'26 & \bfsrcT\ controlled-rollout teacher & \bfsigD\ \bfsigR\ \bfstOn\ progressive/truncated horizons & \bfarchDec\ efficient reasoning OPD \\
Prefix OPD \citep{zhang2026prefixopd} & arXiv'26 & \bfsrcT\ prefix-scoring teacher & \bfsigD\ \bfsigR\ \bfstOn\ truncated prefixes & \bfarchDec\ efficient reasoning OPD \\
DP-OPD \citep{khadem2026dpopd} & arXiv'26 & \bfsrcT\ frozen teacher & \bfsigD\ \bfsigR\ \bfstOn\ DP token targets & \bfarchDec\ privacy-preserving LMs \\
MAD-OPD \citep{wang2026madopd} & arXiv'26 & \bfsrcM\ debating teacher pool & \bfsigD\ \bfsigR\ \bfstOn\ debate-weighted tokens & \bfarchAgent\ agents and code LLMs \\
Draft-OPD \citep{lei2026draftopd} & arXiv'26 & \bfsrcT\ target verifier & \bfsigF\ \bfsigR\ \bfstOn\ draft-error feedback & \bfarchDec\ speculative draft models \\
OmniOPD \citep{zhou2026omniopd} & arXiv'26 & \bfsrcB\ black-box teacher & \bfsigF\ \bfsigR\ \bfstOn\ chunk verification & \bfarchDec\ logit-free reasoning \\
Cross-Tokenizer OPD \citep{niu2026tokenizerbarrier} & arXiv'26 & \bfsrcH\ cross-family teacher & \bfsigD\ \bfsigR\ \bfstOn\ token-mapped distributions & \bfarchDec\ cross-tokenizer LLMs \\
\bottomrule
\end{tabularx}
\vspace{0.35ex}
\begin{tabularx}{\textwidth}{@{}>{\bfseries}p{0.17\textwidth}X@{}}
Source relation &
\bfsrcT\ teacher--student or expert--learner relation;\quad
\bfsrcM\ multiple source models or model zoo;\quad
\bfsrcB\ black-box or API-access source;\quad
\bfsrcS\ self-distillation source;\quad
\bfsrcH\ heterogeneous or cross-modal source--target setting.\\
Behavior signal &
\bfsigD\ output distributions, token probabilities, or logits;\quad
\bfsigX\ demonstrations, responses, rationales, or traces;\quad
\bfsigF\ preferences, rankings, scores, critiques, rewards, or verifier labels;\quad
\bfsigR\ target-induced states, rollouts, or trajectories.\\
State distribution &
\bfstOff\ off-policy fixed behavior data;\quad
\bfstOn\ on-policy supervision on states induced by the target model;\quad
\bfstSemi\ semi-on-policy, cached, or offline-reused on-policy-style supervision.\\
Backbones/settings &
\bfarchEnc\ encoder-only LM;\quad
\bfarchDec\ decoder-only LLM;\quad
\bfarchChat\ chat or instruction-following LLM;\quad
\bfarchMM\ multimodal, speech, or video-language setting;\quad
\bfarchAgent\ imitation-learning or agent-policy setting.\\
Cross-level labels &
\takeB+\takeR\ and \takeR+\takeB\ denote output distillation combined with representation matching; these methods are cross-referenced in Table~\ref{tab:repr_level_fusion_analysis}.\\
Boundary cases &
\textsuperscript{\dag}DAgger is included as the classical on-policy imitation-learning analogue of behavior-level fusion; methods are grouped according to the behavior-level branch in Section~3.3. Labels are descriptive and non-exclusive, since a method may combine distributions, demonstrations, and feedback.
\end{tabularx}
\caption[]{Behavior-level fusion methods compared by source-model relation, behavior signal, state distribution, and evaluated backbones or settings (continued).}
\end{table*}

\section{Model Fusion Benchmarks}

We compare fusion benchmarks by modality, task, heterogeneity, type, and evaluation focus.

\definecolor{BYes}{HTML}{2563EB}
\definecolor{BPartial}{HTML}{D97706}
\definecolor{BNo}{HTML}{9CA3AF}
\definecolor{BVision}{HTML}{0E7490}
\definecolor{BText}{HTML}{059669}
\definecolor{BLLM}{HTML}{7C3AED}
\definecolor{BMLLM}{HTML}{BE185D}

\newcommand{\benchbadge}[2][]{%
  \tikz[baseline=(c.base)]%
  \node[circle, fill=BYes, text=white, font=\scriptsize\bfseries,
  inner sep=0.85pt, minimum size=1.35em, draw=none, #1](c){\strut #2};}

\newcommand{\cmarkb}{\benchbadge[fill=BYes]{Y}}
\newcommand{\pmarkb}{\benchbadge[fill=BPartial]{P}}
\newcommand{\xmarkb}{\benchbadge[fill=BNo]{N}}

\newcommand{\modV}{\benchbadge[fill=BVision]{V}}
\newcommand{\modT}{\benchbadge[fill=BText]{T}}
\newcommand{\modL}{\benchbadge[fill=BLLM]{L}}
\newcommand{\modM}{\benchbadge[fill=BMLLM]{M}}

\begin{table*}[p]
\centering
\scriptsize
\setlength{\tabcolsep}{2.2pt}
\renewcommand{\arraystretch}{1.08}
\begin{tabularx}{\textwidth}{
@{}>{\raggedright\arraybackslash}p{0.24\textwidth}
>{\raggedright\arraybackslash}p{0.10\textwidth}
>{\centering\arraybackslash}p{0.04\textwidth}
>{\centering\arraybackslash}p{0.04\textwidth}
>{\centering\arraybackslash}p{0.04\textwidth}
>{\centering\arraybackslash}p{0.04\textwidth}
>{\centering\arraybackslash}p{0.055\textwidth}
>{\centering\arraybackslash}p{0.055\textwidth}
>{\centering\arraybackslash}p{0.05\textwidth}
>{\raggedright\arraybackslash}p{0.120\textwidth}
>{\raggedright\arraybackslash}X
c@{}}
\toprule
\textbf{Resource} &
\textbf{Venue} &
\textbf{Vision} &
\textbf{Text} &
\textbf{LLM} &
\textbf{MLLM} &
\textbf{Model Pool} &
\textbf{Task Pool} &
\textbf{Hetero.} &
\textbf{Fusion Type} &
\textbf{Evaluation Focus} &
\textbf{Open} \\
\midrule

Realistic Evaluation \citep{tam2024realisticevaluation} &
arXiv'24 &
\cmarkb & \cmarkb & \xmarkb & \xmarkb &
\cmarkb & \cmarkb & \pmarkb &
Vision / text merging &
Acc., compositionality &
\cmarkb \\

Model-GLUE \citep{zhao2024modelglue} &
NeurIPS D\&B'24 &
\xmarkb & \xmarkb & \cmarkb & \xmarkb &
\cmarkb & \cmarkb & \cmarkb &
Heterogeneous LLM merging &
Model selection, aggregation &
\cmarkb \\

MergeKit \citep{goddard2024arcees} &
EMNLP-I'24 &
\xmarkb & \xmarkb & \cmarkb & \xmarkb &
\pmarkb & \pmarkb & \pmarkb &
Recipe-based merging &
Leaderboard perf. &
\cmarkb \\

EMR-Merging \citep{huang2024emrmerging} &
NeurIPS'24 &
\cmarkb & \cmarkb & \pmarkb & \pmarkb &
\cmarkb & \cmarkb & \pmarkb &
Tuning-free merging &
Acc., scalability &
\cmarkb \\

Merging at Scale \citep{khalifa2024ifyou} &
arXiv'24 &
\xmarkb & \xmarkb & \cmarkb & \xmarkb &
\cmarkb & \cmarkb & \xmarkb &
LLM merging &
Scaling, expert count &
\xmarkb \\

H3Fusion \citep{tekin2024h3fusionhelpful} &
EACL'26 &
\xmarkb & \xmarkb & \cmarkb & \xmarkb &
\cmarkb & \cmarkb & \xmarkb &
Alignment merging &
Helpful., honest., harmless. &
\pmarkb \\

SMM-Bench \citep{akizuki2025surrogatebenchmarks} &
AutoML-N'25 &
\xmarkb & \xmarkb & \cmarkb & \xmarkb &
\cmarkb & \cmarkb & \pmarkb &
Surrogate merge search &
Search cost, ranking &
\cmarkb \\

Systematic Study \citep{hitit2025systematicstudy} &
arXiv'25 &
\xmarkb & \xmarkb & \cmarkb & \xmarkb &
\cmarkb & \cmarkb & \xmarkb &
LLM merging study &
Method reliability &
\xmarkb \\

Mergenetic \citep{minut2025mergeneticsimple} &
ACL Demo'25 &
\xmarkb & \xmarkb & \cmarkb & \xmarkb &
\pmarkb & \pmarkb & \pmarkb &
Evolutionary merging &
Fitness, search efficiency &
\cmarkb \\

FusionBench \citep{tang2025fusionbenchcomprehensive} &
JMLR'25 &
\cmarkb & \cmarkb & \cmarkb & \pmarkb &
\cmarkb & \cmarkb & \pmarkb &
Merging / ensemble / mixing &
Acc., robust., OOD &
\cmarkb \\

MergeBench \citep{he2025mergebenchbenchmark} &
NeurIPS D\&B'25 &
\xmarkb & \xmarkb & \cmarkb & \xmarkb &
\cmarkb & \cmarkb & \xmarkb &
Domain LLM merging &
Acc., forgetting, runtime &
\cmarkb \\

OptMerge \citep{wei2025optmergeunifying} &
ICLR'26 &
\xmarkb & \xmarkb & \xmarkb & \cmarkb &
\cmarkb & \cmarkb & \cmarkb &
MLLM merging &
VQA, OCR, grounding &
\cmarkb \\

Merging Scaling Law \citep{wang2025modelmerging} &
ICML'26 &
\xmarkb & \xmarkb & \cmarkb & \xmarkb &
\cmarkb & \cmarkb & \xmarkb &
Large-scale LLM merging &
Scaling law, expert count &
\cmarkb \\

M2RL \citep{wang2026mixmerge} &
arXiv'26 &
\xmarkb & \xmarkb & \cmarkb & \xmarkb &
\cmarkb & \cmarkb & \xmarkb &
RLVR merging / OPD &
Synergy, interference, efficiency &
\cmarkb \\

\bottomrule
\end{tabularx}

\vspace{0.35ex}
\begin{tabularx}{\textwidth}{@{}>{\bfseries}p{0.16\textwidth}X@{}}

Pool definition &
\textit{Model Pool} indicates whether the resource explicitly defines source models, expert models, or fine-tuned checkpoints to be fused;\quad
\textit{Task Pool} indicates whether it defines downstream tasks or domain pools for post-merge evaluation.\\

Heterogeneity &
\textit{Hetero.} indicates whether the resource explicitly evaluates heterogeneous fusion, including cross-family, cross-architecture, cross-modal, or heterogeneous-output-space settings.\\

Open &
\textit{Open} indicates whether the resource provides public code, scripts, model pools, evaluation resources, or reproducible configurations.\\

Symbols &
\cmarkb\ explicit support;\quad
\pmarkb\ partial, implicit, or recipe-dependent support;\quad
\xmarkb\ not covered or not the focus.\\

Boundary cases &
Toolkits and search ecosystems, such as MergeKit \citep{goddard2024arcees} and Mergenetic \citep{minut2025mergeneticsimple}, are included when they provide reusable fusion pipelines or practical evaluation settings. They are not treated as fixed benchmark suites.

\end{tabularx}

\caption{
Representative model fusion benchmarks and related evaluation resources.
Unlike traditional LLM benchmarks that mainly define task instances and metrics, model fusion benchmarks often define source model pools, target tasks, fusion settings, and cost or retention axes.
}
\label{tab:model_fusion_benchmark_analysis}
\end{table*}

\section{Model Fusion Metrics}

Table~\ref{tab:fusion_metrics} compares model fusion metrics and their use conditions.

\begin{table*}[t]
\centering
\small
\setlength{\tabcolsep}{4pt}
\renewcommand{\arraystretch}{1.08}
\begin{tabularx}{\textwidth}{@{}>{\raggedright\arraybackslash}p{0.17\textwidth}>{\raggedright\arraybackslash}p{0.25\textwidth}>{\raggedright\arraybackslash}p{0.31\textwidth}>{\raggedright\arraybackslash}X@{}}
\toprule
\textbf{Dimension} & \textbf{Question} & \textbf{Representative metrics} & \textbf{Use condition} \\
\midrule
Overall quality &
Is the target model strong overall? &
Avg score, mean task score, mean capability score. &
Core metric for most benchmarks. \\
Capability retention &
How much source capability is kept? &
Normalized performance, retention ratio, worst source-task drop. &
Core metric when source tasks or source capabilities are known. \\
Interference and transfer &
Does fusion hurt or help related tasks? &
Local task drop, negative transfer rate, held-out task score. &
Useful when task combinations or held-out settings are defined. \\
Internal alignment &
Are the fusion signals well matched? &
Representation drift, output-distribution gap, calibration error. &
Requires hidden states, logits, or output distributions. \\
Efficiency &
How costly is fusion and use? &
Fusion compute, source-query count, inference latency. &
Needed when comparing practical fusion methods. \\
Safety and risk &
Does fusion keep task constraints? &
Harmful response rate, backdoor attack success rate, privacy leakage. &
Use under a clear safety goal or threat model. \\
\bottomrule
\end{tabularx}
\caption{Common metrics for model fusion. Avg score and normalized performance are the most direct metrics. Other metrics are useful but depend on the setting, access level, or safety goal.}
\label{tab:fusion_metrics}
\end{table*}

\section{Additional Deployment Challenges}
\label{app:continual_large_scale_challenges}

We discuss continual forgetting and scaling beyond Section~\ref{sec:challenges}.

\paragraph{Continual Fusion Can Easily Cause Forgetting.}
In real deployment, the target model may continually absorb new source models, domain updates, or safety patches.
Each fusion step can overwrite earlier knowledge or weaken previously aligned behavior, especially when old training data, source models, or evaluation signals are unavailable.
AIMMerging \citep{feng2025aimmergingadaptive}, NUFILT \citep{qiu2025nullspace}, and K-Merge \citep{shenaj2025kmerge} study continual fusion for language models, but stable long-term fusion remains open.
Behavior-level methods also face forgetting when new skills are learned from source feedback \citep{shenfeld2026selfdistillation}.
Future methods should add skills without retraining while preserving capabilities and safety.

\paragraph{Large-Scale Fusion Remains Underexplored.}
Model fusion can be cheaper than retraining or using all source models at inference time, but large-scale fusion brings new costs.
For parameter-level fusion, MergeKit \citep{goddard2024arcees} makes LLM fusion easier to run.
MergePipe \citep{wang2026mergepipebudget} further shows that expert-parameter I/O and repeated scans become key bottlenecks as the source pool grows.
For method search, FusionBench \citep{tang2025fusionbenchcomprehensive} and MergeBench \citep{he2025mergebenchbenchmark} improve standard comparison, but large models still make candidate evaluation costly.
For behavior-level fusion, source feedback can also be expensive.
Lightning OPD \citep{wu2026lightning} and TIP \citep{xu2026tip} reduce live teacher serving or token-level supervision cost.

\end{document}